\documentclass[12pt,a4paper]{article}
\usepackage{latexsym}
\usepackage{amsmath,amssymb}
\usepackage[dvips]{graphicx}
\usepackage{times}

\begin{document}

\newfont{\Bb}{msbm10 scaled\magstep1}
\newcommand{\bmlambda}{\mbox{\boldmath$\lambda$\unboldmath}}
\newcommand{\bmkappa}{\mbox{\boldmath$\kappa$\unboldmath}}
\newcommand{\bmomega}{\mbox{\boldmath$\omega$\unboldmath}}
\newcommand{\bmupsilon}{\mbox{\boldmath$\upsilon$\unboldmath}}
\newcommand{\bmepsilon}{\mbox{\boldmath$\epsilon$\unboldmath}}
\newcommand{\bmvarepsilon}{\mbox{\boldmath$\varepsilon$\unboldmath}}
\newcommand{\bmvarphi}{\mbox{\boldmath$\varphi$\unboldmath}}
\newcommand{\bmphi}{\mbox{\boldmath$\phi$\unboldmath}}
\newcommand{\bmpsi}{\mbox{\boldmath$\psi$\unboldmath}}
\newcommand{\bmvartheta}{\mbox{\boldmath$\vartheta$\unboldmath}}
\newcommand{\bmtheta}{\mbox{\boldmath$\theta$\unboldmath}}
\newcommand{\bmgamma}{\mbox{\boldmath$\gamma$\unboldmath}}
\newcommand{\bmalpha}{\mbox{\boldmath$\alpha$\unboldmath}}
\newcommand{\bmbeta}{\mbox{\boldmath$\beta$\unboldmath}}
\newcommand{\bmsigma}{\mbox{\boldmath$\sigma$\unboldmath}}
\newcommand{\bmxi}{\mbox{\boldmath$\xi$\unboldmath}}
\newcommand{\bmeta}{\mbox{\boldmath$\eta$\unboldmath}}
\newcommand{\bmnu}{\mbox{\boldmath$\nu$\unboldmath}}
\newcommand{\bmmu}{\mbox{\boldmath$\mu$\unboldmath}}
\newcommand{\bmtau}{\mbox{\boldmath$\tau$\unboldmath}}
\newcommand{\bmpi}{\mbox{\boldmath$\pi$\unboldmath}}

\newcommand{\bmLambda}{\mbox{\boldmath$\Lambda$\unboldmath}}
\newcommand{\bmKappa}{\mbox{\boldmath$\Kappa$\unboldmath}}
\newcommand{\bmOmega}{\mbox{\boldmath$\Omega$\unboldmath}}
\newcommand{\bmUpsilon}{\mbox{\boldmath$\Upsilon$\unboldmath}}
\newcommand{\bmPhi}{\mbox{\boldmath$\Phi$\unboldmath}}
\newcommand{\bmPsi}{\mbox{\boldmath$\Psi$\unboldmath}}
\newcommand{\bmTheta}{\mbox{\boldmath$\Theta$\unboldmath}}
\newcommand{\bmGamma}{\mbox{\boldmath$\Gamma$\unboldmath}}
\newcommand{\bmAlpha}{\mbox{\boldmath$\Alpha$\unboldmath}}
\newcommand{\bmBeta}{\mbox{\boldmath$\Beta$\unboldmath}}
\newcommand{\bmSigma}{\mbox{\boldmath$\Sigma$\unboldmath}}
\newcommand{\bmXi}{\mbox{\boldmath$\Xi$\unboldmath}}
\newcommand{\bmEta}{\mbox{\boldmath$\Eta$\unboldmath}}
\newcommand{\bmNu}{\mbox{\boldmath$\Nu$\unboldmath}}
\newcommand{\bmMu}{\mbox{\boldmath$\Mu$\unboldmath}}
\newcommand{\bmTau}{\mbox{\boldmath$\Tau$\unboldmath}}
\newcommand{\bmPi}{\mbox{\boldmath$\Pi$\unboldmath}}

\begin{center}
{\Large Optimal Design and Optimal Control of Elastic Structures }\\
\vspace{1mm}
{\Large Undergoing Finite Rotations and Elastic Deformations}\\
\vspace{1mm}
 A.\,Ibrahimbegovi\'{c}$^1$, C.\,Knopf-Lenoir$^2$,
A.\,Ku\v{c}erov\'{a}$^1$\, and \, P.\,Villon$^2$
\\
\vspace*{6mm} $^1$Ecole Normale Sup\'{e}rieure - Cachan, LMT, GCE,
61,\,avenue du pr\'esident
 Wilson,\\
94235 Cachan, France, email: {\it ai@lmt.ens-cachan.fr} \\
\vspace{1mm}
$^2$Universit\'{e} de Technologie - Compiegne, Lab. Roberval, GSM,\\
60200 Compiegne, France
\end{center}
\vspace*{5mm}

\section*{Abstract}

In this work we deal with the optimal design and optimal control of structures
undergoing large rotations. In other words, we show how to find the
corresponding initial configuration and the corresponding set of multiple load
parameters in order to recover a desired deformed configuration or some
desirable features of the deformed configuration as specified more precisely by
the objective or cost function. The model problem chosen to illustrate the
proposed optimal design and optimal control methodologies is the one of
geometrically exact beam. First, we present a non-standard formulation of the
optimal design and optimal control problems, relying on the method of Lagrange
multipliers in order to make the mechanics state variables independent from
either design or control variables and thus provide the most general basis for
developing the best possible solution procedure. Two different solution
procedures are then explored, one based on the diffuse approximation of response
function and gradient method and the other one based on genetic algorithm. A
number of numerical examples are given in order to illustrate both the
advantages and potential drawbacks of each of the presented procedures.

\noindent {\bf Keywords: structures, finite rotation,
optimization, control}

\section{Introduction} \label{sec_intro}

Modern structures should often be designed to  withstand very
large displacements and rotations and remain fully operational.
Moreover, the construction phase has also to be mastered and
placed under control, with trying to precisely guide the large
motion of a particular component of the total structural assembly.
Ever increasing demands to achieve more economical design and
construction thus require that the problems of this kind be placed
on a sound theoretical and computational basis, such as the one
explored in this work. Namely, optimization methods can be called
upon to guide the design procedure and achieve desired reduction
of mechanical and/or geometric properties. Similarly, the control
methods are employed to provide an estimate of the loads and the
minimal effort in placing the structure, or its component,
directly into an optimal (desired) shape. Either of these tasks,
optimal design or optimal control, can formally be presented as
the minimization of the chosen cost or objective function
specifying precisely the desired goal. The main difference between
two procedures concerns the choice of the variables defining the
cost function: the design variables are typically related to the
mechanical properties (e.g. Young's modulus) or geometry of the
structure (e.g. particular coordinates in the initial
configuration), whereas the control variables are related to the
actions (e.g. forces) applied on the structure in order to place
it into desired position. Rather then insisting upon this
difference and treating the optimal design and optimal control in
quite different manners (as done in a number of traditional
expositions on the subject), we focus in this work on their common
features which allow a unified presentation of these two problems
and the development of a novel solution procedure applicable to
both problems. The latter implies that the nonlinear mechanics
model under consideration of geometrically exact beam has to be
placed on the central stage and one should show how to fully
master the variation in chosen system properties or loads in order
to achieve the optimal goal. The main contributions put forward in
presenting this unconventional approach can be stated as follows:

\begin{enumerate}
\item [i)] We first present the theoretical framework for treating
nonlinear structural mechanics and optimization and/or control as
a coupled (nonlinear) problem; In any such problem of optimization
or control, the nonlinear mechanics equilibrium equations are
reduced to a mere constraint with respect to the admissibility of
a given state of the structure, i.e. its displacements and
rotations. By using the classical method of Lagrange multipliers
(e.g. see \cite{Luenberger:1984} or \cite{Strang:1986} the
mechanics equilibrium equations can be promoted from constraint to
a one the governing equations to be solved in a coupled problem of
this kind, and the intrinsic dependence on state variables
(displacements and rotations) with respect to optimal design or
control variables can be eliminated turning all variables into the
independent variables. For clarity, this idea is also developed
within the framework of a discrete, finite-element-based
approximation, thus providing the finite element model including
the degrees of freedom pertinent not only to displacements and
rotations, but also to optimal design and/or control variables. A
detailed development is presented for the chosen model problem of
3D geometrically exact beam (e.g. see \cite{Simo:1986} or
\cite{Ibrahimbegovic:1995}).  We note in passing that the proposed
approach is quite opposite from the traditional ones (e.g. see
\cite{Kleiber:1997} for a very recent review), where the two
fields directly concerned by the coupled problems of this kind,
nonlinear mechanics on one side and optimal design or control on
the other, are studied and developed separately and resulting
solution procedures for one or another are applied in a sequential
manner. One typically employs two different computer programs, one
for mechanics and another for optimization or control (e.g. the
toolbox of {\it MATLAB}), so that the communication requirements
are reduced to a bear minimum: so-called sensitivity (e.g.
\cite{Tortorelli:1994} or \cite{Rousselet:1992}) for optimization
code, and design or control variables for the finite element code
for mechanics. It is clear that such a traditional approach to
analysis and design or control will largely sacrifice the
computational efficiency for the cases of practical interest where
both the cost function and mechanics problem are nonlinear and
require iterative procedures to be solved.

\item [ii)] The second aspects which is elaborated upon in this
work pertains to an alternative method for solving a coupled
problem for analysis and either design or control, where two
sub-problems, once brought up to the same level by the Lagrange
multiplier method, are solved simultaneously. In other words, once
the interdependence of the state variables, i.e. the displacements
and rotations on one side and the design or control variables or
another, is no longer enforced, one can iterate simultaneously to
solve for all of them. In particular, the sensitivity analysis
needs no longer be performed separately, but it is naturally
integrated as a part of the simultaneous iterative procedure. It
is important to note that the iterative intermediate values are no
longer consistent with the equilibrium equations constraint,
except at the convergence, where basically the same solution is
obtained as for the standard sequential solution procedure but
with a (significantly) reduced number of iterations. In other
words the simultaneous and sequential solution procedures will
always yield the same result providing the solution is unique. The
problems of this kind concern the optimization and/or control of
geometrically nonlinear response of structures where no
bifurcation phenomena occur. Solving the nonlinear problems in
structural mechanics can be quite a demanding task for the
structures where the stiffness may differ significantly in
different deformation modes (i.e. beam bending versus stretching).
Adding the control or optimization problem on top makes such a
nonlinear problem so much more challenging. Different
modifications of the basic solution procedure are developed and
tested, including genetic algorithms \cite{Hrstka:2003:Elsevier}
to explore the initial phase of the solution procedure, gradient
based acceleration near solution points and response surface for a
part of the solution constructed by diffuse approximation (e.g.
see \cite{Breitkopf:2002}).
\end{enumerate}

The outline of the paper is as follows. In the next section we briefly present
the chosen model of geometrically exact 3D beam, capable of representing
very large displacements and rotations. The theoretical formulation of the
optimal control and optimal design of the chosen mechanical model is presented
in Section \ref{sec_cop}, along with their discrete
approximations constructed by the finite element method given in Section
\ref{sec_mef}. The proposed solution procedure is described in detail in Section
\ref{sec_sol_proc}. Several numerical examples are presented in Section
\ref{sec_examples} in order to illustrate the performance of various algorithms
used in computations. Some closing remarks are stated in Section
\ref{sec_conclusions}.


\section{Model problem: geometrically exact 3D beam} \label{sec_mod_prob}

In this section, we briefly review the governing equations of the
chosen model problem of a structure undergoing large rotations, a
3D curved beam. For a more thorough discussion of the chosen model
we refer to \cite{Simo:1986}, \cite{Simo:1992} or
\cite{Ibrahimbegovic:Frey:1995}, among others. We assume that the
initial configuration of the beam is internal force free and that
it can be described by a 3D position vector $ {\bmvarphi }_{0}$(s)
identifying the position of each point of the neutral fiber (an
inextensible fiber in pure bending) and the corresponding
placement of the cross-section of the rod, which is carried out by
choosing a local ortho-normal triad of vectors. The vector triad
of this kind can be obtained by simply rotating the global triad
by an orthogonal tensor.
For a usual
choice of so-called normal coordinates with the first vector of
the triad being orthogonal to the cross-section and the remaining
two placed in the plane of the cross section, this orthogonal
tensor becomes a known function of the initial configuration, $
{\bf \Lambda }_{0} (s)$. For the case of a curved beam studied
here, '$s$' is chosen as the arc length parameter.

By applying the external loading $ {\bf f}(t)$, parameterized by
pseudo-time 't' (where 'pseudo' implies that the inertia effects
are neglected) we obtain the beam deformed configuration defined
by the position vector $ { \bmvarphi }(s,t)$ and the orthogonal
tensor $ {\bf \Lambda }(s,t)$. The latter is an accordance with
the usual kinematic hypothesis that the cross-section of the beam
would not deform, which, along with the hypothesis that the first
vector in the triad remains orthogonal to it (with other two
within the plane of the cross-section) fully determines $ { \bf
\Lambda }(s,t)$. See Figure 1 where the initial and deformed
configuration of the beam are presented.

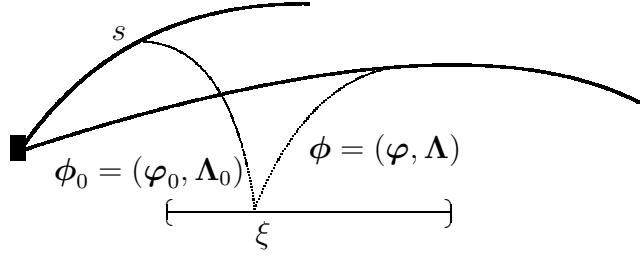
\begin{figure}
\centering
\unitlength=1mm \special{em:linewidth 0.4pt} \linethickness{0.4pt}
\begin{picture}(98.33,36.33)
\put(15.75,14.50){\rule{2.00\unitlength}{3.33\unitlength}}
\thicklines \bezier{850}(17.33,16.33)(30.67,35.67)(55.00,35.33)
\bezier{1000}(17.67,16.00)(74.00,34.67)(98.33,22.33) \thinlines
 \put(36.33,7.67){\oval(0.00,3.33)[l]}
\put(73.67,7.50){\oval(0.00,3.67)[r]}
\put(36.33,7.67){\line(1,0){37.33}}
\bezier{100}(48.00,7.67)(45.00,29.67)(33.33,30.33)
\bezier{80}(47.67,7.67)(55.00,26.00)(67.33,27.00)
\put(30.00,31.50){\makebox(0,0)[cc]{$s$}}
\put(49.00,4.67){\makebox(0,0)[cc]{$\xi$}}
\put(34.00,13.00){\makebox(0,0)[cc]{$ {\bmphi}_0 = (
 {\bmvarphi}_0,  {\bf \Lambda}_0)$}}
\put(65.00,16.00){\makebox(0,0)[cc]{$ {\bmphi} = (
 {\bmvarphi},  {\bf \Lambda})$}}
\end{picture}
\caption{Initial and deformed configuration of the 3D geometrically exact
beam.}
\label{fig_defbeam}
\end{figure}

In short, one can state that the configuration space of the
described model of 3D beam can be written as

$\qquad $%
\begin{equation}
{\cal C}:=\{ {\bmphi}= (
 {\bmvarphi}, {\bf \Lambda } )\, | \,
 {\bmvarphi}\in \mathbb{R}^{3}, {\bf \Lambda }\in
SO(3)\}
\end{equation}
\par\noindent
where $\mathbb{R}^{3}$ and $SO(3)$ are spaces of 3D vectors and
special orthogonal tensors, respectively.

The main difficulty in numerical solution of the structural mechanics problems
featuring the beams of this kind stems from the presence of $SO(3)$
group in its configuration space (e.g. see \cite{Argyris:1982},
\cite{Ibrahimbegovic:Frey:1995}, \cite{Ibrahimbegovic:1997} for a more
thorough discussion of these issues). In short, in performing the standard task
of computing the virtual work principle or the consistent linearization, where a
small rotation described by a skew-symmetric tensor $\delta
 {\bf \Theta }\in so(3)$, ought to be superposed on a large
rotation described by an orthogonal tensor $ {\bf \Lambda }\in
SO(3)$ one must first make use of exponential mapping

\begin{equation}
 {\bf \Lambda }_{\varepsilon }= {\bf \Lambda }\exp
[\varepsilon \, \delta {\bf
 \Theta }]\ ; \quad
\exp [\delta  {\bf \Theta }]=\cos \delta \theta\,  {\bf
I}+\frac{\sin \delta \theta }{\delta \theta }\delta  {\bf \Theta}
+ \frac{1-\cos \delta \theta }{\delta \theta ^{2}} \delta  {
\bmtheta }\otimes  \delta  { \bmtheta }
\end{equation}
\par\noindent
where $\delta  {\bmtheta }$ is the axial vector of the
skew-symmetric tensor $\delta  {\bf \Theta }$ i.e. $\delta { \bf
\Theta} \; {\bf v}=\delta  {\bmtheta} \times  {\bf v},$ $ \quad
\forall  {\bf v}\in \mathbb{ R}^{3} $.

The complexity of the last expression is in sharp contrast, with respect to a
simple additive update of virtual displacement field $\delta
 {\bmvarphi }\in \mathbb{R}^{3}$\, when superposed on the deformed
configuration $ {\bmvarphi } \in \mathbb{R}^{3}$

\begin{equation}
 {\bmvarphi }_{\varepsilon} =  {\bmvarphi } +
\varepsilon\,  \delta  {\bmvarphi }
\end{equation}

The results in (2) and (3) can be presented in an equivalent form
by stating that the tangent space of the chosen beam model is defined by

\begin{equation}
T{\cal C}:=\{\delta  {\bmphi }:=(\delta  {\bmvarphi },\delta
 {\bmtheta})\; | \delta  {\bmvarphi }\in \mathbb{R}^{3},
\delta  {\bmtheta} \in \mathbb{R}^{3}\}
\end{equation}

The strain measures employed in this beam theory (e.g.
\cite{Ibrahimbegovic:1995}) can be written in direct tensor notation form as

\begin{equation}
 {\bmepsilon}=  {\bf \Lambda} {\bmvarphi
}^{\prime }
\end{equation}
\par\noindent
for the axial and shear strains, and

\begin{equation}
 {\bf \Omega }=  {\bf \Lambda} ^{T}  {\bf \Lambda}
^{\prime }\ ; \quad
 {\bf \Omega} {\bf v}= {\bmomega} \times
 {\bf v}\ ;\ \forall  {\bf v}\in \mathbb{R}^{3}
\end{equation}
\par\noindent
for bending and torsional strains. In (5), (6) and subsequent
equations we denote with superposed prime the derivative with
respect to arc-length coordinate in the initial configuration, i.e.

\begin{equation}
\frac{\partial }{\partial s}(\cdot )=(\cdot )^{\prime }
\end{equation}

We consider the simplest case of linear elastic material model for
the beam which allows us to express the constitutive equations in
terms of stress resultants as

\begin{equation}
 {\bf n}=  {\bf C}( {\bmepsilon}-
 {\bmepsilon}_{0})\ ; \quad  {\bf C} = diag (EA,GA,GA)
\end{equation}

\begin{equation}
 {\bf m}= {\bf D}(  {\bmomega}-
 {\bmomega}_{0})\ ; \quad   {\bf D} = diag (GJ,EI,EI)
\end{equation}

For illustration, we also consider the simplest case of a circular cross
section, with section diameter '$d$', and
\begin{equation}
A = \frac{d^{2}\pi }{4}\ ; \quad I = \frac{d^{4}\pi }{64}\ ; \quad
J = \frac{d^{4}\pi }{ 32}
\end{equation}
\par\noindent
as the section area, moment of inertia and polar moment.

In order to complete the description of the chosen beam model we
state the equilibrium equations in the weak form as

\begin{equation}
G( {\bmphi }; \delta  { \bmphi }):=\int (\delta  { \bmepsilon}
\cdot  {\bf n}+ \delta  { \bmomega }\cdot  {\bf m})\, ds -
G_{ext}(\delta  {  \bmphi })=0
\end{equation}
\par\noindent
where $G_{ext}(\delta  { \bmphi })$ is the external virtual work
and $\delta  { \bmepsilon }$\ and $\delta  {\bmomega }$\ are the
virtual strains. The latter can be obtained as the G\^{a}teaux
derivative of the real strains in (5) and (6), by taking the
results in (2) and (3) into consideration. In particular, this
leads to

\begin{eqnarray}
\delta \widehat{ {\bmepsilon }}( {\bmphi };\delta  { \bmphi } )
&=& D_{\phi }[\widehat{ {\bmepsilon }}( {\bmphi })] \notag
\\ &=&\frac{d}{d\varepsilon }[{\bf \Lambda}_{t,\varepsilon }^{T}\bmvarphi
_{t,\varepsilon }^{\prime }]\mid _{\varepsilon =0} \\ &=&
 {\bf \Lambda }^{T}\delta  {\bmvarphi }^{\prime} +
 {\bmepsilon } \times \delta \theta  \notag
\end{eqnarray}
\par\noindent
and

\begin{eqnarray}
\delta \widehat{\bmOmega }( {\bmphi };\delta  { \bmphi }) &=&
D_{\phi }[\widehat{\bmOmega }(\bmphi_{t,\varepsilon })] \notag
\\ &=&\frac{d}{d\varepsilon }[{\bf \Lambda}_{t,\varepsilon }^{T}{\bf \Lambda}
_{t,\varepsilon }^{\prime }]\mid _{\varepsilon =0} \\ &=&\delta
 {\bf \Theta }^{\prime }+ \delta  {\bf \Theta }^{T}\bmOmega
+\bmOmega \delta  {\bf \Theta}  \notag
\end{eqnarray}
\par\noindent
which can also be written in an equivalent form in terms of the
corresponding axial vectors

\begin{equation}
\delta  {\bmomega} =\delta  {\bmtheta}^{\prime }+  {\bmomega}
\times \delta  {\bmtheta}
\end{equation}


For the external load which derives from a given potential, such that
\begin{equation}
G_{ext}(\delta\phi):=D_{\phi}\Pi_{ext}(\bmphi)
\end{equation}
one can also define the governing equilibrium equation of the chosen model
problem of the geometrically exact elastic beam from the principle of minimum of
the total potential energy defined according to
\begin{equation}
\Pi ( {\bmphi}):=\int_{l}\frac{1}{2}\{( {\bmepsilon }- {
\bmepsilon }_{0})\cdot  {\bf n}+( {\bmomega }- {\bmomega
}_{0})\cdot   {\bf m} \} \, ds - \Pi_{ext} ( {\bmphi}) \rightarrow
\min
\end{equation}
\par\noindent
which implies that
\begin{equation}
\Pi ({\phi})=\underset{\forall  {\bmphi}^{\ast }}{\min }\Pi (
{\bmphi}^{\ast })\Longrightarrow \left\{
\begin{array}{l}
D_{\phi }[\Pi ( {\bmphi})] \equiv G(  {\bmphi};\delta
 {\bmphi} )= 0 \\ D_{\phi }[D_{\phi }\Pi
( {\bmphi})] \equiv \underbrace{D_{\phi}G(\bmphi; \delta
\bmphi)}_{\delta \bmphi \cdot {\bf K} \delta \bmphi}>0
\end{array}
\right.
\end{equation}

In (17) above we denoted by {\bf K} the second variation of the
total potential energy or the tangent operator, which is obtained
by the consistent linearization procedure (e.g. see
\cite{Marsden:1983}). For pertinent details of the consistent
linearization we refer to \cite{Ibrahimbegovic:1995} for
quaternion parameterization of finite rotations or
\cite{Ibrahimbegovic:Frey:1995} for rotation vector-like
parameters for finite rotations.

An alternative case of external loading where the potential may not be defined
and where the weak form in (11) is rather the starting point of the solution
procedure, can also be encountered in applications. For example, in the case of
the follower force ${\bf p}_0$ and follower moment ${\bf l_0}$ which follow the
motion of a particular cross-section of the beam at node '$a$', we can express
their contribution to virtual work according to
\begin{equation}
G_{ext}({\bf \Lambda_a};\delta \bmvarphi_a, \delta
\bmtheta_a):=\delta \bmvarphi_a \cdot \underbrace{{\bf \Lambda}
{\bf p}_0}_{{\bf p}} + \delta \bmtheta_a \cdot \underbrace{{\bf
\Lambda} {\bf l}_0}_{{\bf l}}
\end{equation}

The follower force and moment will also contribute to the tangent operator
according to
\begin{equation}
D_{\phi} G_{ext} ({\bf \Lambda}; \delta \bmvarphi_a, \delta
\bmtheta_a) \cdot \left( \begin{array}{c}
\Delta \bmvarphi_a \\
\Delta \bmtheta_a
\end{array}
\right) = \delta \bmvarphi_a \cdot {\bf P} \Delta \bmtheta_a +
\delta \bmtheta_a \cdot {\bf L} \Delta \bmtheta_a
\end{equation}
\par\noindent
where $ {\bf P} {\bf v} = {\bf p} \times {\bf v}; {\bf L} {\bf v}
= {\bf l} \times {\bf v}; \forall {\bf v} \in {\mathbb R}^3$. This
contribution should also be taken into account when computing the
solution to (11) and trying to ensure the quadratic convergence
rate.


\section{Coupled optimality problem} \label{sec_cop}

The presented beam model provides an excellent basis to master the optimization
problem as well as the control problem of geometrically nonlinear elastic
structures. Although the former deals with geometric characteristics of the beam
and the latter with the external loading sequence, the two problems can be
formulated and solved in quite an equivalent manner, as shown next.


\subsection{Optimal design} \label{subsec_design}


The optimal design problem addressed herein pertains to selecting
the desired features of the mechanical model by leaving the free
choice of the geometric properties of the beam model, e.g. the
thickness variation or the chosen initial shape. This task is
often referred to as the shape optimization. From the mathematical
standpoint the shape optimization can be formulated as the problem
of minimization of so-called objective or cost function
$J(\cdot)$, specifying the desired features. The latter is
considered as a functional which depends not only on mechanics
state variables $\bmphi$ but also on design variables ${\bf d}$
(e.g. the beam thickness or its shape).

The shape optimization procedure is interpreted herein as
minimization of so-called cost or objective functional ${\hat
J}(\cdot $), which can be written as

\begin{equation}
\hat J( \hat{\bmphi}(d),d)=\underset{ G(  \hat{\bmphi}^{\ast
}(d^{\ast}) ; \cdot )=0}{\min } \hat J( \hat{\bmphi}(d^{\ast}),
d^{\ast }) \label{eq_3.1.1}
\end{equation}

Contrary to the minimization of the total potential energy
functional in (17), not all mechanical and design variables are
admissible candidates, but only those for which the weak form of
the equilibrium equations is satisfied. In other words, we now need
to deal with a constrained minimization problem.

The classical shape optimization procedure of solving this
constrained minimization problem is carried out in a sequential
manner, where for each iterative value of design variables
$d^{(i)}$, a new iterative procedure must be completed leading to
$ {\bmphi}(d^{(i)})$ verifying the equilibrium equations. The
considerable computational cost of such a procedure, most of it
waisted on iterating to convergence on equilibrium equations even
for non-converged values of design variables, can be drastically
reduced by formulating the minimization problem in
(\ref{eq_3.1.1}) using the method of Lagrange multipliers with

\begin{equation}
\underset{\forall \lambda }{\max }\underset{\forall (
{\bmphi}^{\ast },\,d^{\ast })}{\min } L ( {\bmphi }^{\ast }, {\bf
d}^{\ast }; \ {\bmlambda })\ ; \quad L( {\bmphi }^{\ast }, {\bf
d}^{\ast }; {\bmlambda })= J( { \bmphi }^{\ast }, {\bf d}^{\ast })
+ G( {\bmphi }^{\ast }, {\bf d}^{\ast }; {\bmlambda })
\label{eq_3.1.2}
\end{equation}

In (\ref{eq_3.1.2}) above ${\bmlambda }$ are the
Lagrange multipliers inserted into the weak form of equilibrium
equations instead of virtual displacements and rotations. In
accordance to the results presented in the previous section we can write
explicitly

\begin{equation}
G( \bmphi, d; \bmlambda )= \int_l \{ \bmlambda^{\prime}_{\varphi}
\cdot \bmLambda {\bf n} + \bmlambda_{\theta} \cdot ({\bf E}^T {\bf
n} + \bmOmega^T {\bf m}) +\bmlambda^{\prime}_{\theta} \cdot {\bf
m} \} \mbox{d}s - G_{ext}(\bmlambda) \label{eq_3.1.3}
\end{equation}

The main difference of (\ref{eq_3.1.2}) with respect to
constrained minimization problem in (\ref{eq_3.1.1}) pertains to
the fact that state variables $ {\bmphi}$ and design variables
${\bf d}$ are now considered independent and they can be iterated
upon (and solved for) simultaneously.

The Kuhn-Tucker optimality condition (e.g. \cite{Luenberger:1984})
associated with the minimization problem in (18) can be written as

\begin{equation}
 {\bf 0} = D_{\phi} \left[ L  (  {\bmphi }, {d};
{\bmlambda }) \right] = D_{\phi }\left[ J (   {\bmphi}, {d}
 )\right] + D_{\phi }\left[ G  (  {\bmphi },{d};
 {\bmlambda } )\right]
\label{eq_3.1.4}
\end{equation}
\par\noindent
with the explicit form of the last term which can be written as
\begin{equation}
\begin{array}{l}
 D_{\phi }\left[ G( {\bmphi }, {d};
{\bmlambda })\right] \cdot \delta  {\bmphi}  = \int_l \left\{
\bmlambda^{\prime}_{\varphi} \cdot ( \bmLambda {\bf C} \bmLambda^T
\delta \bmvarphi^{\prime} + \bmLambda {\bf CE} \delta \bmtheta )
\right. \\ \left. + \bmlambda_{\theta} \cdot [ {\bf E}^T {\bf C}
\bmLambda^T \delta \bmvarphi^{\prime} + ( {\bf E}^T {\bf CE} +
\bmOmega^T {\bf D} \bmOmega ) \delta \bmtheta + \bmOmega^T {\bf D}
\delta \bmtheta^{\prime} ] + \bmlambda^{\prime}_{\theta} \cdot (
{\bf D} \bmOmega \delta
\bmtheta + {\bf D} \delta \bmtheta^{\prime} ) \right\} \, ds \\
+ \int_l \left\{ \bmlambda^{\prime}_{\varphi} \cdot ( \bmLambda
{\bf N}^T \delta \bmtheta ) + \bmlambda_{\theta} \cdot  [ {\bf N}
\bmLambda^T \delta \bmvarphi^{\prime} + \left[ \bmXi \left(
\bmepsilon \times {\bf n} \right) + \bmXi \left( \bmomega \times
{\bf m} \right) \right] \delta \bmtheta + {\bf M} \delta
\bmtheta^{\prime} ] \right\} \, ds  \label{eq_3.1.5}
\end{array}
\end{equation}
\par\noindent
where we denoted $ {\bf \Xi}( {\bf a} \times  {\bf b})= ( {\bf a}
\otimes  {\bf b}) -  ( {\bf a} \cdot  {\bf b})  {\bf I}$, as well
as $ {\bf M}  {\bf v} =  {\bf m} \times  {\bf v}$ and $ {\bf N}
{\bf v} =  {\bf n} \times {\bf v}$, $\forall {\bf v} \in
\mathbb{R}^3$; Moreover, we have
\begin{equation}
0 = D_{d}\left[ L(\cdot )\right]  = D_{d}J(\cdot ) + D_{d} G(\cdot)
\label{eq_3.1.6}
\end{equation}
\par\noindent
 where
\begin{equation}
D_{d}\left[ G (\cdot )\right] \cdot \delta {\bf d} =
\underset{l}{\int }\left\{ \bmlambda_{\varphi}^{\prime} \cdot
\bmLambda \frac{\partial  {\bf n}}{\partial {\bf d}} +
\bmlambda_{\theta} \cdot \left( {\bf E}^T \frac{\partial {\bf
n}}{\partial {\bf d}} + \bmOmega^T \frac{\partial  {\bf
m}}{\partial {\bf d}} \right) + \bmlambda_{\theta}^{\prime} \cdot
\frac{\partial  {\bf m}}{\partial {\bf d}} \right\} \cdot \delta
{\bf d} \mbox{d}s \label{eq_3.1.7}
\end{equation}
\par\noindent

Finally, we also have
\begin{equation}
0 = D_{\lambda }[ L( \cdot) ] \cdot \delta  {\bmlambda} =\int_l
\left\{ \delta \bmlambda^{\prime}_{\varphi} \cdot \bmLambda {\bf
n} + \delta \bmlambda_{\theta} \cdot \left( {\bf E}^T {\bf n} +
\bmOmega^T {\bf m} \right) + \delta \bmlambda_{\theta}^{\prime}
\cdot {\bf m} \right\} \mbox{d} s \label{eq_3.1.8}
\end{equation}

For illustration we can consider further that the diameter of a
circular cross-section is chosen as the design variable, which allows us to
express explicitly the result in (23) as

\begin{align}
\frac{\partial  {\bf n}}{\partial d}& =\frac{\partial
 {\bf C}}{\partial d}(\varepsilon -\varepsilon _{0})\ ;\
  \frac{\partial
 {\bf C}}{\partial d}=diag[E\frac{\partial A}{\partial d}, G\frac{\partial
A}{\partial d},G\frac{\partial A}{\partial d}]\ ; \ A = \frac{d^2 \pi}{4}; \label{eq_3.1.9} \\
\frac{\partial
 {\bf m}}{\partial d}& =\frac{\partial  {\bf D}}{\partial
d}(\omega -\omega _{0})\ ;\  \frac{\partial
 {\bf D}}{\partial
d}=diag[G\frac{\partial J}{\partial d},E\frac{\partial I}{\partial d},E\frac{%
\partial I}{\partial d}] \ ; \
   J = 2I = \frac{d^4
\pi}{32} \label{eq_3.1.10}
\end{align}
\par\noindent
In order to provide a similar explicit result for directional
derivative of the cost function, we consider a simple choice given
as the beam mass (or equivalently its volume for a constant density),

\begin{equation}
J( {\bmphi},d) \equiv V =\underset{l}{\int}A\,  \mbox{d}s \ ;
\quad A = \frac{d^2 \pi}{4} \label{eq_3.1.11}
\end{equation}

In such a case the contribution of the cost function to the
Kuhn-Tucker optimality conditions can be written as

\begin{align}
D_{\phi } J( {\bmphi},d)\cdot \delta  {\bmphi }& =0 \notag
\\
D_{d} J ( {\bmphi },d)\cdot \delta d& =\underset{l}{\int }\frac{%
\partial A}{\partial d} \delta d\, ds\ ; \quad \frac{\partial A}{\partial d} =
\frac{d \pi}{2} \label{eq_3.1.12}
\\
D_{\lambda }J( {\bmphi },d)\cdot \delta  {\bmlambda }& =0 \notag
\end{align}

We can also consider a more complex case of practical interest where the shape
optimization is carried out with respect to the beam axis form in the initial
configuration. A reference configuration is selected in such a case (see Figure
1) and the design variable is given in terms of the position vector
describing the beam initial configuration with respect to this
reference configuration $ {\bf d} \equiv   {\bmvarphi }_{0} \left(
\xi \right)$. The cost function in (25) can now be described as

\begin{equation}
J ( {\bmphi }, {\bf d}) :=\underset{l}{\int } A\, ds =
\underset{\xi_{1}}{\overset{\xi_{2}}{\int }} A\, j(\xi )\, d\xi\ ;\qquad j(\xi
)=\left\| \frac{\partial {\bf d}(\xi )}{\partial \xi }\right\|
\label{eq_3.1.13}
\end{equation}

In this case all the integrals in (\ref{eq_3.1.4}) to
(\ref{eq_3.1.8}) must be recomputed with the same change of
variables like the one presented in (\ref{eq_3.1.13}) above
typical of the isoparametric parent element mapping (e.g. see
Zienkiewicz and Taylor \cite{Zienkiewicz:2000}). We also note in
passing that the derivatives with respect to arc-length coordinate
ought to be computed by making use of the chain rule

\begin{equation}
\frac{d}{ds}(\cdot ) = \frac{1}{j(\xi )}\frac{\partial }{\partial
\xi }(\cdot )
\label{eq_3.1.14}
\end{equation}

For example, the contribution of the cost function to the Kuhn-Tucker
optimality conditions for such a choice of design variables can be
written as

\begin{equation}
D_{d}\left[ J( {\bmphi }, {\bf d})\right] \cdot \delta  {\bf d}=%
\underset{\xi_{1}}{\overset{\xi_{2}}{\int }} A \frac{1}{j(\xi )} {\bf d}%
(\xi )\cdot \frac{\partial  {\bf d}(\xi )}{\partial \xi }d\xi
\label{eq_3.1.15}
\end{equation}


\subsection{Optimal control} \label{subsec_control}

The optimal control problem studied herein concerns the
quasi-static external loading sequence which is chosen to bring
the structure directly towards an optimal or desired final state,
which may involve large displacements and rotations. More
precisely, we study the mechanics problems where introducing the
pseudo-time parameter 't' to describe a particular loading program
is not enough and one also needs to employ the control variables
$\bmnu$. The latter contributes towards the work of external
forces, which can be written as
\begin{equation}
G^{ext} ( \bmnu; \delta \bmphi ) := \int_l \{ \delta \bmphi \cdot
{\bf F}_0 \bmnu \} ds \label{eq_3.2.1}
\end{equation}
where ${\bf F}_0$ contains the (fixed) distribution of the external loading to
be scaled by the chosen control.

Optimal control can be presented in the following form: find the
value of the control variables $\bmnu$, such that the final value
of the state variables $\bmphi$ be as close as possible to the
desired optimal (fixed) values $\bmphi_d$; This can be formulated
in terms of the constrained minimization of the chosen cost
function $J(\hat{\bmphi}(\bmnu),\bmnu)$ which can be written as
\begin{equation}
\hat{J}(\hat{\bmphi}(\bmnu), \bmnu) = \min_{
G(\hat{\bmphi}(\nu^{\ast});\cdot)=0}\hat
J(\hat{\bmphi}(\bmnu^{\ast}), \bmnu^{\ast})\biggr|_{\hat \phi_d}
\label{eq_3.2.2}
\end{equation}
where the role of the constraint, as indicated  by the last
expression, is to fix the values of the state variables with
respect to the chosen value of the control through the given set
of equilibrium equations. In other words, for a given control we
will finally obtain the configuration $\bmphi$ the same as the
desired state $\bmphi_d$ if the latter verifies equilibrium
equations, or, in the opposite case simply the solution closest to
$\bmphi_d$ which also verifies the equilibrium equations.

In order to remove this constraint and to be able to consider the
state variables independently from the control variables, we
resort to the classical method of Lagrange multipliers; Namely, by
introducing the Lagrange multipliers $\bmlambda$ we can rewrite
the optimal control problem by making use of the Lagrangian
functional $L(\cdot)$ which allows to obtain the corresponding
form of the unconstrained optimization problem.
\begin{equation}
\max_{\forall \bmlambda} \min_{\forall (\bmphi,\bmnu)} L (\bmphi,
\bmnu, \bmlambda)\biggr|_{\bar{\phi}_d}\ ; \quad L(\bmphi, \bmnu,
\bmlambda) =
 J(\bmphi, \bmnu) + G(\bmphi, \bmnu; \bmlambda)
\label{eq_3.2.3}
\end{equation}

One can readily obtain the Kuhn-Tucker optimality conditions for
this problem according to
\begin{eqnarray}
0 &=& {\partial L \over \partial {\bmlambda}} \cdot \delta
{\bmlambda} := G(\bmphi, \bmnu; \delta\bmlambda)
\label{eq_3.2.4-1}
\\
 0 &=& {\partial L
\over
\partial \bmphi} \cdot \delta \bmphi := {\partial J(\cdot) \over
\partial \bmphi} \cdot \delta {\bmphi} + {\partial G(\bmphi;\bmlambda)
\over \partial \bmphi} \cdot \delta {\bmphi} \label{eq_3.2.4-2}
\end{eqnarray}
and
\begin{equation}
 0 = {\partial L \over \partial \bmnu} \delta \bmnu
:=   {\partial J \over \partial \bmnu} \delta \bmnu + {\partial
G^{ext} \over \partial \bmnu} \delta \bmnu \label{eq_3.2.4-3}
\end{equation}
The first of these equations is precisely the weak form of
equilibrium equation, whereas the second two will provide the
basis for computing the control $\bmnu$ and the Lagrange
multipliers $\bmlambda$. The explicit form of the equilibrium
equation is similar to the one in (11) only with the external load
term (\ref{eq_3.2.1}), but with the variation of the Lagrange
multiplier $\delta \bmlambda$ replacing the virtual displacement
$\delta \bmvarphi$. For writing the explicit form of other two
Kuhn-Tucker equations we choose a particular form of the objective
function such that
\begin{equation}
J(\bmphi, {\bmnu} ) = \int_l \left\{ \frac{1}{2} \| \bmphi -
\bar{\bmphi}_d \|^2 + \frac{1}{2} \alpha \| {\bmnu} \|^2 \right\}
ds \label{eq_3.2.5}
\end{equation}
where $\bar{\bmphi}_d$ is the desired beam shape and $\alpha$ is a scalar
parameter specifying the weighted contribution of the chosen control. With
such a choice we seek to minimize the "distance" between the desired shape and
the computed admissible shape (the one which satisfies equilibrium) as well as
the force or control needed to achieve that state. The explicit form of the
first term in (\ref{eq_3.2.4-2}) and (\ref{eq_3.2.4-3}) can thus be written as
\begin{equation}
{\partial J \over \partial \bmphi} \cdot \delta \bmphi = \int_l
\left\{ \delta \bmphi \cdot \left( {\bmphi} - \bar{\bmphi}_d
\right) \right\} ds
\end{equation}
and
\begin{equation}
{\partial J \over \partial \bmnu} \cdot \delta {\bmnu} = \int_l
\left\{ \delta {\bmnu} \cdot \left( \alpha {\bmnu} \right)
\right\} ds \label{eq_3.2.6}
\end{equation}
Explicit result for the second term of Kuhn-Tucker equations in
(\ref{eq_3.2.4-2}) is identical to the one in (\ref{eq_3.1.5}) potentially
modified according to (19) for the case of the follower load. Finally, the
second term of the Kuhn-Tucker equation in (\ref{eq_3.2.4-3}) can be written as
\begin{equation}
{\partial G^{ext} \over \partial {\bmnu} } \cdot \delta \bmnu =
\int_l \left\{ \delta \bmnu \cdot {\bf F}_0^T \bmlambda \right\}
ds \label{eq_3.2.7.}
\end{equation}
which concludes with the description of all the problem ingredients.


\section{Finite element discrete approximations} \label{sec_mef}


In this section we discuss several important aspects of numerical
implementation of the presented theory for analysis and design and
related issues which arise in numerical simulations.

The analysis part of the problem, i.e. the state variables are
represented by using the standard isoparametric finite element
approximations (e.g. see \cite{Zienkiewicz:2000}). In particular,
this implies that the element initial configuration is represented
with respect to its parent element
placed in the natural coordinate space, corresponding to a fixed interval, $%
-1=\xi _{1}\leq \xi \leq \xi _{2}=+1$, by using
\begin{equation}
 {\bmvarphi }_{0}\left( s \right) \equiv  {\bf x}\left( \xi \right) =%
\underset{a=1}{\overset{n_{en}}{\sum }}N_{a}\left( \xi \right)
 {\bf x}_{a}
 \label{eq_4.1}
\end{equation}
\par\noindent
In (\ref{eq_4.1}) above  $ {\bf x}\left( \xi \right) $\ is the
position vector field with respect to the reference configuration,
$ {\bf x}_{a}$\ are nodal values of an element with $n_{en}$ nodes
and $N_{a}\left( \xi \right) $\ are the corresponding shape
functions. The latter can easily be constructed for beams by using
the Lagrange polynomials, which for an element with $n_{en}$ nodes
can be written by using the product of monomial expressions
\begin{equation}
N_{a}\left( \xi \right) =\underset{b=1,b\neq a}{\overset{n_{en}}{\prod }}%
\frac{\xi -\xi _{b}}{\xi _{a}-\xi _{b}}
\end{equation}
\par\noindent
where $\xi _{a}$ , $a\in \left[ 1,n_{en}\right] $ are the nodal
values of natural coordinates.

With isoparametric interpolations one chooses the same shape
functions in order to approximate the element displacement field,
which allows us to construct the finite element representation of
the element deformed configuration as
\begin{equation}
 {\bmvarphi }\left( \xi \right) =\underset{a=1}{\overset{n_{en}}{\sum }}%
N_{a}\left( \xi \right)  {\bmvarphi }_{a}
\end{equation}
\par\noindent
where $ {\bmvarphi}_{a}$\ are the nodal values of the position
vector in the deformed configuration. The virtual and incremental
displacement field are also represented by isoparametric
finite element interpolations
\begin{equation}
\delta  {\bmvarphi }\left( \xi \right) =\underset{a=1}{\overset{n_{en}%
}{\sum }}N_{a}\left( \xi \right) \delta  {\bmvarphi }_{a}\ ; \quad
\Delta  {\bmvarphi }\left( \xi \right) =\underset{a=1}{\overset{n_{en}%
}{\sum }}N_{a}\left( \xi \right) \Delta  {\bmvarphi }_{a}
\label{eq_4.4}
\end{equation}
The latter enables that a new (iterative) guess for the deformed
configuration be easily \ obtained with the corresponding additive
updates of the nodal values

\begin{equation*}
 {\bmvarphi}_{a} \longleftarrow
 {\bmvarphi}_{a}+\Delta {\bmvarphi }_{a}
\end{equation*}

The finite element approximation of the incremental displacement
field in (\ref{eq_4.4}) above, where at each point $\xi \in \left[ \xi
_{1},\xi _{2}\right] $ the corresponding value is a linear
combination of the nodal values, are referred to as the continuum
consistent (e.g. see \cite{Ibrahimbegovic:1994,Ibrahimbegovic:1995}), since they
allow to commute the finite element interpolation and the consistent
linearization of nonlinear problem (e.g. \cite{Marsden:1983}).
Therefore, we also choose the isoparametric
interpolations for virtual and incremental rotation field with

\begin{equation}
\delta \bmtheta \left( \xi \right) =\underset{a=1}{\overset{n_{en}}{\sum }}%
N_{a}\left( \xi \right) \delta \bmtheta _{a}\ ; \quad
\Delta \bmtheta \left( \xi \right) =\underset{a=1}{\overset{n_{en}}{\sum }}%
N_{a}\left( \xi \right) \Delta \bmtheta _{a}
\end{equation}
so that the commutativity of the finite element discretization and
consistent linearization would also apply to the rotational state
variables. The only difference from the displacement field
concerns the multiplicative updates of the rotation parameters,
which can be written for any nodal point $'a'$ as

\begin{equation}
 {\bf \bmLambda}_{a}\leftarrow {\bf \bmLambda}_{a}\exp\left[
\Delta\bmtheta_{a}\right]
\end{equation}

In the combined analysis and design procedure proposed herein one must
also interpolate the Lagrange multipliers, which is also
done by using the isoparametric interpolations according to

\begin{equation}
 {\bmlambda }\left( \xi \right) =\underset{a=1}{\overset{n_{en}}{\sum }}%
N_{a}\left( \xi \right)  {\bmlambda }_{a}\Longleftrightarrow
\left\{
\begin{array}{c}
 {\bmlambda}_{\varphi} \left( \xi \right) =\underset{a=1}{\overset{n_{en}}{\sum
}} N_{a}\left( \xi \right)  {\bmlambda }_{\varphi_a} \\
 {\bmlambda }_{\theta} \left( \xi \right) =\underset{a=1}{\overset{n_{en}}{\sum
}} N_{a}\left( \xi \right)  {\bmlambda }_{\theta_a}
\end{array}
\right.
\end{equation}

The corresponding integrals appearing in governing Lagrangian
functional in (\ref{eq_3.1.2}) or Kuhn-Tucker optimality conditions in
(\ref{eq_3.1.4}), (\ref{eq_3.1.6}) and (\ref{eq_3.1.8}) are computed by
numerical integration (e.g. Gauss quadrature, see \cite{Zienkiewicz:2000}). To
illustrate these ideas we further state a single element contribution to the
analysis part of the governing Lagrangian functional in (\ref{eq_3.1.3}) given
in the discrete approximation setting by

\begin{equation*}
G ( {\bmphi}_{a}, {\bf d},\lambda _{a}):=\left(
\begin{array}{c}
 {\bmlambda}_{\varphi_a} \\  {\bmlambda}_{\theta_a}
\end{array}
\right) \overset{n_{ip}}{\underset{l=1}{\sum }}\left[
\begin{array}{ccc}
\frac{dN_{a}(\xi _{l})}{ds}   {\bf I} & {\bf 0 } & {\bf 0}
\\ \mathbf{0} & N_{a}(\xi _{l}) {\bf I} & \frac{dN_{a}(\xi_l)}{ds}  {\bf I}
\end{array}
\right]^T
\end{equation*}
\begin{equation}
\left[
\begin{array}{ccc}
 {\bf \Lambda }(\xi _{l}) &  {\bf E}(\xi _{l}) &
 {\bf 0}
\\  {\bf 0} &  {\bf \Omega }(\xi_{l}) &  {\bf I}
\end{array}
\right] ^{T}\left(
\begin{array}{c}
 {\bf n}(\xi _{l}) \\  {\bf m}(\xi _{l})
\end{array}
\right) j(\xi )w_{l}-G_{ext}( {\bmlambda} (\xi _{l}))
\end{equation}
\par\noindent
where $\xi _{l}$\ and $w_{l}$\ are the abscissas and weights of the chosen
numerical integration rule (e.g. see \cite{Zienkiewicz:2000}) and $n_{ip} $ is
the total number of integration points for a single element.

In order to complete the discretization procedure one must also
specify the interpolations of the design variables. If the latter
is the thickness or the diameter for the chosen case of a circular
cross-section, or the element nodal coordinates, it is possible to
use again the isoparametric finite element approximations.
However, the best results are obtained by reducing the number of
design variables as opposed to those chosen at the element level,
by using the concept of a design element. This implies increasing
the degree of polynomial by employing, for example, B\'{e}zier and
B-spline curves for representation of beam shape and reducing
significantly the number of design parameters (e.g. see
\cite{Kegl:2000} for a detailed discussion of these ideas). What
is important to note from the standpoint of the simultaneous
solution procedure presented further on is that the design
variable at any point is given as a linear combination of the
design element interpolation parameters

\begin{equation}
 {\bf d}(\xi_{l})=\underset{a=1}{\overset{n_{dn}}{\sum}}B_{a}\left( \xi
_{l}\right)  {\bf d}_{a}
\end{equation}

Consequently, the finite or rather design element discretization
and consistent linearization will commute again. This observation was
already made earlier for linear analysis problem by
Chenais and Knopf-Lenoir \cite{Chenais:1989}. With these results on hand we can
write the discrete approximation of equilibrium equations as


\begin{eqnarray}
\lefteqn{D_{\lambda} L ( \bmphi, {\bf d}; \bmlambda ) := 0 \mapsto}  \nonumber \\
\lefteqn{\underset{e=1}{\overset{n_{el}}{\mbox{\Huge A}}} \left\{
{\bf r}_a^{\lambda, e} = \underset{e=1}{\overset{n_{int}}{\sum}} \left[
\begin{array}{cccc}
\frac{d N_a (\xi_e)}{ds} {\bf 1} & {\bf 0} & {\bf 0} \\
{\bf 0} & N_a (\xi_e) {\bf 1} & \frac{d N_a (\xi_e)}{ds} {\bf 1}
\end{array}
\right]^T \right. }  \nonumber \\
 & & \qquad \qquad \quad \, \left. \left[
\begin{array}{ccc}
\bmLambda^T(\xi_e) & {\bf E}(\xi_e) & {\bf 0} \\
{\bf 0} & \bmOmega(\xi_e) & {\bf I}
\end{array}
\right]^T \left(
\begin{array}{c}
{\bf n}(\xi_e) \\
{\bf m}(\xi_e)
\end{array}
\right) j(\xi_e) - {\bf f}_a^{ext}
\right\} = {\bf 0}
\end{eqnarray}
where 'A' denotes the finite element assembly operator.

By the same token the discrete approximation of the
Kuhn-Tucker optimality conditions in (\ref{eq_3.1.4}) can be presented as
\begin{eqnarray}
\lefteqn{D_{\phi} L ( \bmphi, {\bf d}; \bmlambda ) := 0 \mapsto} \nonumber \\
\lefteqn{\underset{e=1}{\overset{n_{el}}{\mbox{\Huge A}}} \left\{ {\bf
r}_a^{\phi, e} = \underset{e=1}{\overset{n_{int}}{\sum}} \left[ \left[
\begin{array}{cccc}
\frac{d N_a (\xi_e)}{ds} {\bf 1} & {\bf 0} & {\bf 0} \\
{\bf 0} & N_a (\xi_e) {\bf 1} & \frac{d N_a (\xi_e)}{ds} {\bf 1}
\end{array}
\right]^T \left[
\begin{array}{ccc}
\bmLambda^T(\xi_e) & {\bf E}(\xi_e) & {\bf 0} \\
{\bf 0} & \bmOmega(\xi_e) & {\bf I}
\end{array}
\right]^T \right. \right. }  \nonumber \\
 & & \qquad \qquad \qquad \left[
\begin{array}{cc}
{\bf C} & {\bf 0} \\
{\bf 0} & {\bf D}
\end{array}
\right] \left[
\begin{array}{ccc}
\bmLambda^T(\xi_e) & {\bf E}(\xi_e) & {\bf 0} \\
{\bf 0} & \bmOmega(\xi_e) & {\bf I}
\end{array}
\right] + \nonumber \\
 & & \qquad \qquad \qquad + \left. \left[
\begin{array}{ccc}
{\bf 0} & \bmLambda^T(\xi_e)N^T(\xi_e) & {\bf 0} \\
N(\xi_e)\bmLambda^T(\xi_e) & {\bf \Xi}( {\bmepsilon} \times {\bf
n})+
{\bf \Xi}( {\bmomega} \times  {\bf m}) &  {\bf M} \\
{\bf 0} & {\bf 0} & {\bf 0}
\end{array}
\right] \right] \nonumber \\
 & & \qquad \quad \ \left. \sum_{b=1}^{n_{en}}\left[
\begin{array}{ccc}
\frac{d N_b (\xi_e)}{ds} {\bf 1} & {\bf 0} & {\bf 0} \\
{\bf 0} & N_b (\xi_e) {\bf 1} & \frac{d N (\xi_e)}{ds} {\bf 1}
\end{array}
\right]  \left(
\begin{array}{c} \bmlambda_{\varphi b} \\ \bmlambda_{\theta b}
\end{array}
\right) j(\xi_e) w_l
\right\} = {\bf 0}
\end{eqnarray}

Similarly, for the simple choice of the objective function in (\ref{eq_3.1.11})
the discrete approximation of the optimality condition in (\ref{eq_3.1.6}) can
be written as
\begin{eqnarray}
\lefteqn{D_d L ( \bmphi, {\bf d}; \bmlambda ) := 0 \mapsto} \nonumber \\
\lefteqn{\underset{e=1}{\overset{n_{el}}{\mbox{\Huge A}}} \left\{ {\bf r}_a^{d,
e} = \underset{e=1}{\overset{n_{int}}{\sum}} \left[ \left[
\begin{array}{cccc}
\frac{d N_a (\xi_e)}{ds} {\bf 1} & {\bf 0} & {\bf 0} \\
{\bf 0} & N_a (\xi_e) {\bf 1} & \frac{d N_a (\xi_e)}{ds} {\bf 1}
\end{array}
\right]^T \left[
\begin{array}{ccc}
\bmLambda^T(\xi_e) & {\bf E}(\xi_e) & {\bf 0} \\
{\bf 0} & \bmOmega(\xi_e) & {\bf I}
\end{array}
\right]^T  \left(
\begin{array}{c}
\frac{\partial {\bf n}(\xi_e)}{\partial d} \\
\frac{\partial {\bf m}(\xi_e)}{\partial d}
\end{array}
\right) \right. \right. } \nonumber \\
 & & \, \left. \left.  \sum_{b=1}^{n_{dn}} B_b(\xi_e) {\bf
d}_b j(\xi_e) w_l \right]^T + \frac{\partial A (\xi_e)}{\partial d}
\sum_{b=1}^{n_{dn}} B_b(\xi_e) {\bf d}_b j(\xi_e) w_l \right\} = {\bf 0}
\end{eqnarray}

In summary, the discrete approximation of the optimal design
problem reduces to the following set of nonlinear algebraic
equations where the unknowns are the nodal values of displacements
and rotations, the corresponding values of Lagrange multipliers
and the chosen values of thickness parameters which support
B\'ezier curve thickness interpolations,
\begin{equation}
{\bf r}(\bmphi, {\bf d}; \bmlambda) := \left(
\begin{array}{rcl}
{\bf r}^{\lambda} &:=& {\bf f}^{int} - {\bf f}^{ext} \\
{\bf r}^{\phi} &:=& {\bf K} \bmlambda \\
{\bf r}^d &:=& \frac{\partial {{\bf f}^{int}}^T}{\partial {\bf d}} +
\frac{\partial V}{\partial {\bf d}}
\end{array}
\right) = {\bf 0}
\label{eq_4.13}
\end{equation}

Similar procedure leads to the discrete approximation of the
optimal control problem. In fact, the first of the nonlinear
algebraic equations in (54) is only slightly modified to include
the control variables in accordance with the expression in
(\ref{eq_3.2.1})
\begin{equation}
{\bf r}^{\lambda} := {\bf f}^{int} - {\bf F_0}{\bmnu} = {\bf 0}
\label{eq_4.14}
\end{equation}
where ${\bmnu}$ are the chosen control parameters.

For the particular choice of the objective function of the control problem given
in (\ref{eq_3.2.5}), we can further obtain the discrete approximation of the
optimality condition in (\ref{eq_3.2.4-2}) according to
\begin{eqnarray}
D_{\phi}L(\bmphi, \bmnu, \bmlambda) = 0 \quad \Rightarrow \quad
{\bf r}^{\phi}:=\bmphi - \bar{\bmphi}_d + {\bf K}\bmlambda = 0
\label{eq_4.15}
\end{eqnarray}

The discrete approximation of the last optimality condition in
(\ref{eq_3.2.4-3}) can also be written explicitly as
\begin{equation}
D_{\nu} L(\bmphi, \bmnu, \bmlambda) = 0 \quad \Rightarrow \quad
{\bf r}^{\nu}:=\alpha {\bmnu} - {\bf F}_0^T \bmlambda = 0
\label{eq_4.16}
\end{equation}

We note in passing that the last two optimality conditions can be combined to
eliminate the Lagrange multipliers leading to
\begin{equation}
\alpha {\bmnu} + {\bf F}_0^T {\bf K}^{-1} \bmphi = {\bf F}_0^T
{\bf K}^{-1} \bar{\bmphi}_d \label{eq_4.17}
\end{equation}

The last result can be combined with the equilibrium equations in
(\ref{eq_4.14}) providing a reduced set of equations with nodal
displacements, rotations and control variables as the only
unknowns. This kind of form is fully equivalent to the arc-length
solution procedure, which is used for solving the nonlinear
mechanics problems in the presence of the critical points (e.g.
see \cite{Riks:1972}, \cite{Ceisfield:1979} or \cite{Ramm:1988},
among others). One can recognize that (\ref{eq_4.14}) is only a
particular choice of the supplementary condition which is used to
stabilize the system.


\section{Solution procedure} \label{sec_sol_proc}

Two novel solution procedures are developed for solving this class of problems
of optimal design and optimal control as described next.

\subsection{Diffuse approximation based gradient methods} \label{subsec_diffuse}

The first solution procedure is a sequential one, where one first
computes grid values of the cost function and then carry out the
optimization procedure by employing the approximate values
interpolated from the grid. It is important to note that all the
grid values provide the design or control variables along with the
corresponding mechanical state variables of displacements and
rotations which must satisfy the weak form of equilibrium
equation. In other to ensure this requirement, for any grid value
of design or control variables one also has to solve associated
nonlinear problem in structural mechanics.

The main goal of the subsequent procedure is to avoid solving
these nonlinear mechanics problems for other but grid values, and
simply assume that the interpolated values of the cost function
will be "sufficiently" admissible with respect to satisfying the
equilibrium equations. Having relaxed the equilibrium
admissibility requirements we can pick any convenient
approximation of the cost function, which will simplify the
subsequent computation of the optimal value and thus make it much
more efficient. These interpolated values of the cost function can
be visualized as a surface (yet referred to as the {\it response
surface}) trying to approximate sufficiently well the true cost
function. The particular method which is used to construct the
response surface of this kind is the method of diffuse
approximations (see \cite{Nayroles:2002} or \cite{Breitkopf:2002}).
By employing the diffuse approximations the approximate value of
the cost function $\hat J_{appr}$ is constructed as the following
quadratic form
\begin{equation} \hat J_{appr}({\bf x}) = c +
{\bf x}^T{\bf b} + \frac{1}{2}{\bf x}^T{\bf H x} \label{eq_5.1.1}
\end{equation}
where $c$ is a constant reference value, ${\bf b} = ( b_i ) \, ;
\, b_i = \frac{\partial \hat J_{appr}}{\partial { x}_i}$ is the
gradient and ${\bf H} = [ H_{ij} ] \, ; \, H_{ij} =
\frac{\partial^2 \hat J_{appr}}{\partial{ x}_i
\partial x_j}$ is the Hessian of the approximate cost function. In
(\ref{eq_5.1.1}) above variables ${\bf x}$ should be replaced by
either design variables ${\bf d}$ for the case of an optimal
design problem or by control variables ${\bmnu}$ in the case when
we deal with an optimal control problem.

We further elaborate on this idea for a simple case where only 2
design or control variables are used, such that ${\bf x} = (x_1,
x_2)^T$. For computational proposes in such a case one uses the
polynomial approximation typical of diffuse approximation (see
\cite{Breitkopf:2002}) employing the chosen quadratic polynomial
basis ${\bf p}({\bf x})$  and a particular point dependent
coefficient values ${\bf a}({\bf x})$
\begin{equation}
J_{appr}({\bf x}) = {\bf p}^T ({\bf x}) {\bf a}({\bf x}) \ ; \
{\bf p}^T({\bf x}) = [ 1, x_1, x_2, x_1^2, x_1 x_2, x_2^2 ] \ ; \
{\bf a} ({\bf x}) = [a_1({\bf x}), a_2({\bf x}), \ldots , a_6({\bf
x})]
\end{equation}
\par\noindent
By comparing the last two expressions one can easily conclude that
\begin{equation}
c = a_1 \ ; \ {\bf b} = \left( \begin{matrix} a_2 \\ a_3
\end{matrix} \right) \ ; \ {\bf H} = \left[ \begin{matrix}
\begin{matrix} a_4 \\ a_5 \end{matrix} \begin{matrix} a_5 \\ a_6
\end{matrix} \end{matrix} \right]
\end{equation}
\par\noindent
The approximation of this kind is further fitted to the known,
grid values of the cost function; $J({\bf x}_i), i=1,2,\ldots, n$,
trying to achieve that the point-dependent coefficients remain
smooth when passing from one sub-domain to another. This can be
stated as the following minimization problem:
\begin{equation}
{\bf a}({\bf x}) = arg \, \min_{\forall a^*} f ({\bf a}^*,{\bf x})
 \ ; \ f ({\bf a}^*,{\bf x}) :=
\frac{1}{2} \sum_{i=1}^n W ({\bf x}, {\bf x}_i) \left[ J({\bf
x}_i) - {\bf p}^T({\bf x}_i){\bf a}^* \right]^2 \label{eq_5.1.2}
\end{equation}
\par\noindent
where $W({\bf x}, {\bf x}_i)$ are the weighting functions
associated with a particular data point ${\bf x}$, which are
constructed by using a window function $\rho(\cdot)$ based on
cubic splines according to
\begin{equation}
W({\bf x}, {\bf x}_i) =  \rho \left(\frac{\parallel {\bf x} - {\bf
x}_i
\parallel}{r({\bf x})}\right) \ ;  \rho (s) := 1 - 3 s^2 + 2 s^3 \ ; \
r({\bf x}) = \max_k [ dist ({\bf x}, {\bf x}_k) ]
\label{eq5.1.2a}
\end{equation}
\par\noindent
with ${\bf x}_k({\bf x}) , k = 1, \ldots n+1 \ (=7$ for the
present 2-component case) as the closest grid nodes of the given
point ${\bf x}$. We can see that the weighting functions
$W(\cdot)$ in
 (\ref{eq_5.1.2}) and (\ref{eq5.1.2a}) above take a unit value at any of the closest grid nodes
 ${\bf x}_i$ and
vanish outside the given domain of influence. While the former
assures the continuity of the coefficients ${\bf a}({\bf x})$, the
latter ensures that the approximation remains local in character.
Similar construction can be carried out for higher order problems,
which requires an increased number of closest neighbors in the
list.

By keeping the chosen point ${\bf x}$ fixed and considering the
coefficients ${\bf a}$ of diffuse approximation as constants, the
minimization of $f(\cdot)$ amounts to using the pseudo-derivative
of diffuse approximation (see \cite{Nayroles:2002}) in order to
 compute ${\bf x}$ yielding the
minimum of $J_{app}({\bf x})$ according to
\begin{equation}
0 = \sum_{i=1}^n {\bf p}({\bf x}_i) \, W ({\bf x},{\bf x}_i) ( J_i
- {\bf p}^T ({\bf x}_i) {\bf a} )
\end{equation}
\par\noindent
which allows us to write a set of linear equations
\begin{eqnarray}
  {\bf a}({\bf x}) = ({\bf P} {\bf W P}^T)^{-1}{\bf P W j} & & \nonumber \\
  {\bf P} = [{\bf p}({\bf x}_1), {\bf p}({\bf x}_2) \ldots {\bf p}({\bf x}_n)]\,
 ;\, {\bf j} = \left( J({\bf x}_1), J({\bf x}_2), \ldots, J({\bf x}_n) \right) & & \\
  {\bf W} = diag \left( W({\bf
x}_1,{\bf x}), W({\bf x}_2,{\bf x}), \ldots, W({\bf x}_n,{\bf x})
\right) \nonumber
\end{eqnarray}
\noindent We note in passing that the computed minimum value of
$\hat J_{appr}$ does not necessarily provide the minimum of the
true cost function, which also has to satisfy the equilibrium
equations; however, for a number of applications this solution can
be quite acceptable. If the latter is not sufficient, we ought
explore an alternative solution procedure capable of providing the
rigorously admissible value of any computed minima of the cost
function, by carrying out the simultaneous solution of the cost
function minimization and the equilibrium equations. The proposed
procedure is based on genetic algorithm as described next.


\subsection{Genetic algorithm based method} \label{subsec_gen_alg}

Genetic algorithms belong to the most popular
optimization methods nowadays. They follow up an analogy of processes
that run in nature within the evolution of living
organisms over a period of many millions of years. Unlike the
classic gradient optimization methods, genetic algorithms operate on
so called population which is a set of possible solutions, applying the genetic
operators, such as cross-over, mutation and selection. The
principles of genetic algorithm was first proposed by Holland
\cite{Holland:1975}. Ever since, the genetic algorithms have reached wide
application domain (e.g. see the books of Goldberg \cite{Goldberg:1989}
and Michalewicz \cite{Michalewicz:1992} for extensive review).

Genetic algorithms in the original form operate on population of
so-called chromosomes. These are binary strings which represent
possible solutions in a certain way. In engineering problems we
are usually working with real variables, which in the kind of
applications described further are optimized values of load
control or design variables. Adaptation of the genetic algorithm
idea to this problem is made possible by Storn
\cite{Storn:1996:NAPHIS} by considering the chromosomes as vectors
instead of binary strings and using differential operators which
can affect the distance between the chromosomes.

In this work we employ an improved version of this kind of algorithm referred to
as 'Simplified Atavistic Differential Evolution', or SADE (\cite{SADE-WWW}). It
is shown \cite{Hrstka:2003:Elsevier, Leps:2003:Elsevier} that the algorithm is
well suitable for dealing with a fairly large number of variables which is of
particular interest for the problem on hands. The SADE algorithm is designed to
explore all possible minima and thus find the global minimum, even for the case
where the cost function may have very steep gradients and isolated peak values.
Only a short description of the corresponding procedure is given subsequently
(see \cite{Hrstka:2003:Elsevier} for a more elaborate presentation).

In the tradition of evolutionary methods, the first step is
to~generate a starting generation of chromosomes by choosing the
random values of all state variables along with those for control
or design. Subsequently we repeat until convergence the cycles
containing: the creation of~a~new generation of chromosomes, by
mutation, local mutation or cross-over and the evaluation and
selection, which reduces the actual number of~chromosomes to~the
initial number.

In the computations to follow we will work with the population of
'$10 \times n$' chromosomes, where '$n$' is the total number of
unknowns in the problem. This population can evolve through the
following operations;
 Mutation: let ${\bf x}_i(g)$ be the {\em i}-th chromosome in a generation~{\em
g}, \begin{equation}
{\bf x}_i(g)=(x_{i1}(g),x_{i2}(g),...,x_{in}(g)),
\end{equation}
where $n$ is the number of variables of the cost function.
If a~certain chromosome ${\bf x}_i(g)$ is chosen to be mutated, a random
chromosome $RP$ is generated from function domain and a new one ${\bf x}_k(g+1)$
is computed using the following relation:
\begin{equation}
{\bf x}_k(g+1)={\bf x}_i(g)+MR(RP-{\bf x}_i(g)).
\end{equation}
Parameter $MR$ is the constant of algorithm equal to $0.5$. Number
of new chromosomes created by the operator 'mutation' is defined
by 'radioactivity', which is another parameter of the algorithm,
with a constant value set to $0.1$ for all our calculations. If a
certain chromosome is chosen to be locally mutated, all its
coordinates have to be altered by a random value from a given
(usually very small) range. Its aim is to~utilize the near
neighborhood of~existing chromosomes and~make search for~improved
solutions faster. It is useful only for cost functions with steep
gradients in the case, where near the optimal function value, a
small change in value of variable may introduce a large change in
function value. The aim of~the cross operator is to~create as many
new chromosomes as there were in~the last generation. The operator
creates new chromosome ${\bf x}_i(g+1)$ according to the following
sequential scheme: choose randomly two chromosomes ${\bf x}_q(g)$
and ${\bf x}_r(g)$, compute their difference vector, multiply it
by~a~coefficient \verb"CR" and add it to~the third, also randomly
selected chromosome ${\bf x}_p(g)$, i.e,
\begin{equation}
{\bf x}_i(g+1)={\bf x}_p(g)+CR({\bf x}_q(g)-{\bf x}_r(g)).
\label{eq_cross}
\end{equation}

Every component exceeding the defined interval is changed to the
appropriate boundary value of domain. The parameter $CR$ has
probably the most important effect on~the algorithm's behavior; it
seems that if the speed up of~the convergence is needed, this
parameter should be set to~some lower value (from~$0.05$
to~$0.1$). In the opposite case, the higher values of~this
parameter could improve the ability to~solve for any local
minimum. In our computations this value was set to~$0.3$.

Selection represents the kernel of~the genetics
algorithm. The goal is to~provide a~progressive improvement of~the whole
population, which is achieved by~reducing the number of~''living" chromosomes
together with conservation of~better ones. Modified tournament strategy is used
for~this purpose: two chromosomes are chosen randomly from a~population, they
are compared and~the worse of~them is cast off. This conserves
population diversity thanks to~a~good  chance of~survival even for~badly
performing chromosomes.

It was observed that the SADE algorithm can be quite inefficient in the later
stage of the analysis, where the solution with all or a large number of
components is almost converged. For that reason we have tried to further improve
the performance by forcing the algorithm to stick better to these converged
values. In particular we have experimented with a modified form of the cross
operator, which contrary to the one in (\ref{eq_cross}) will produce the new
chromosome by building on top of the best possible previous value according to
\begin{equation}
{\bf x}_i(g+1)=max({\bf x}_q(g);{\bf x}_r(g))+SG.CR({\bf x}_q(g)-{\bf x}_r(g)).
\label{eq_cross_grade}
\end{equation}
where $SG$ is the sign change parameter which is supposed to get
the correct orientation of the increase with respect to the
gradient. Moreover, the parameter $CR$ is no more constant, but
its values are chosen randomly from interval $(0,CL)$, where $CL$
is new parameter of the algorithm, with a smaller influence at its
behavior than $CR$. Operator local mutation is now switched off
and parameter $MR$ in operator mutation is also no more constant,
but chosen randomly from interval $(0,1)$. This algorithm
modification is subsequently referred to as GRADE.


\section{Numerical examples} \label{sec_examples}

In this section we present several illustrative examples dealing
with the coupled problems of mechanics and either optimal control
or optimal design. The computations are carried out by using a
mechanics model of 2D geometrically exact beam (see
\cite{Ibrahimbegovic:1993} or \cite{Ibrahimbegovic:1999})
developed either within the {\it MATLAB} environment for diffuse
approximation based solution procedure or within the $C^{++}$ SADE
computer code for genetic algorithms.


\subsection{Optimal control of a cantilever structure in the form of letter T}

\begin{figure}[htb]
\vspace{5mm}
\begin{center}
\includegraphics[width=100mm,keepaspectratio]{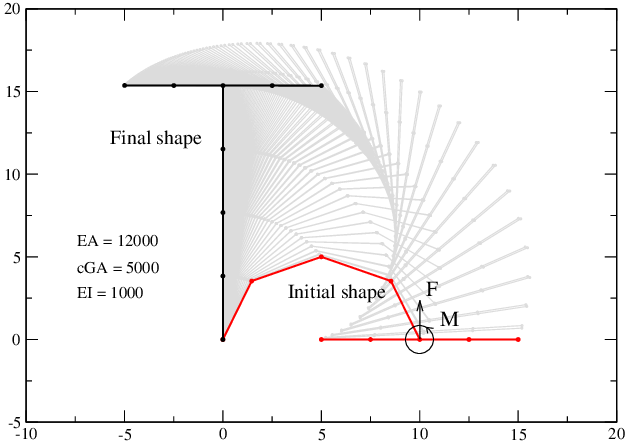}
\caption{T letter cantilever: Initial, final and intermediate
configurations}
\label{tecko1}
\end{center}
\end{figure}

In this example we study the optimal control problem of deploying
initially curved cantilever beam in the final configuration which
takes the form of letter T. See Figure \ref{tecko1} for initial
and final configurations indicated by thick lines and a number of
intermediate deformed configurations indicated by thin lines. The
chosen geometric and material properties are as follows: The
diameter of the circular curved part and the length of the flat
part of the cantilever are both equal to $10$; the beam
cross-section is a unit square; the chosen values of Young's and
shear moduli are $12000$ and $6000$, respectively.

The deployment is carried out by applying a vertical force $F$ and a moment $M$
at the end of the curved part of the cantilever. In other words the chosen
control is represented by a vector $\bmnu = (F, M)^T$. The desired shape of the
cantilever which takes the form of letter T corresponds to the values of force
$F=40$ and moment $M=205$. The optimal control problem is then defined as
follows. The objective or cost function is defined by imposing the desired shape
only on displacement degrees of freedom with
\begin{equation}
J(\bmnu) = \frac{1}{2} \int_L \|\hat{\bmphi}(\bmnu)-\bmphi^d\|^2 \mbox{d}s \\
\label{eq_contr}
\end{equation}
which is further recast in the discrete approximation setting as
\begin{equation}
J^h(\bmnu) =  \frac{1}{4} \sum_{e=0}^{n_{el}} \sum_{a=1}^2 l^e \left( {\bf
u}_a^e(\bmnu) - {\bf u}_a^d \right)^T \left( {\bf u}_a^e(\bmnu) - {\bf u}_a^d
\right)
\label{eq_contr_discr}
\end{equation}
where ${\bf u}_a^e(\bmnu)$ are computed and ${\bf u}_a^d$ are
desired nodal displacements. We note in passing that no condition
is imposed through the cost function on either rotational degrees
of freedom or control vector, which nevertheless introduces no
difficulties in solving this problem.

\begin{figure*}[!ht]
\centering
\begin{tabular}{l@{\hspace{2mm}}l}
\vspace{3mm}
\includegraphics*[width=6.5cm]{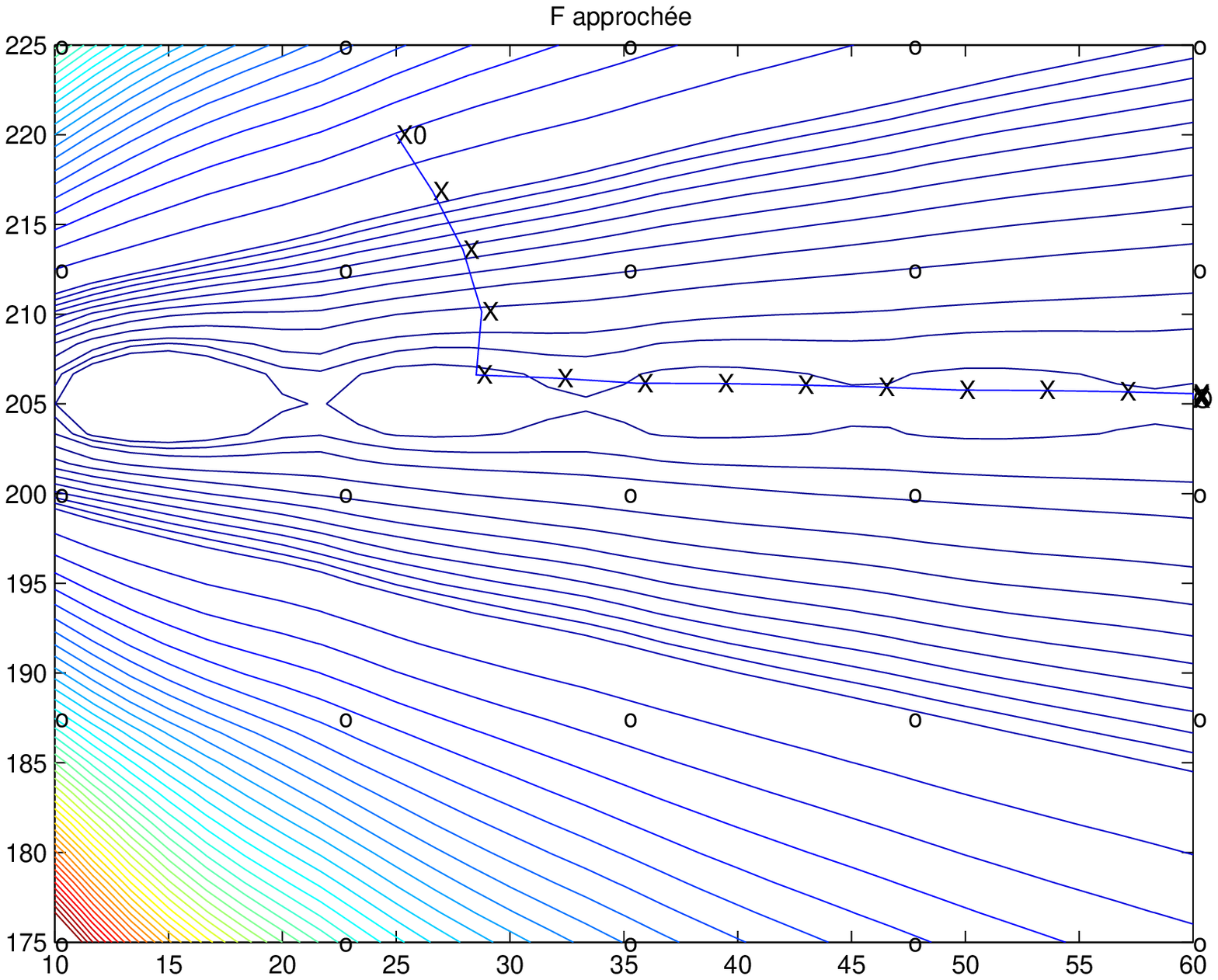} &
\includegraphics*[width=6.5cm]{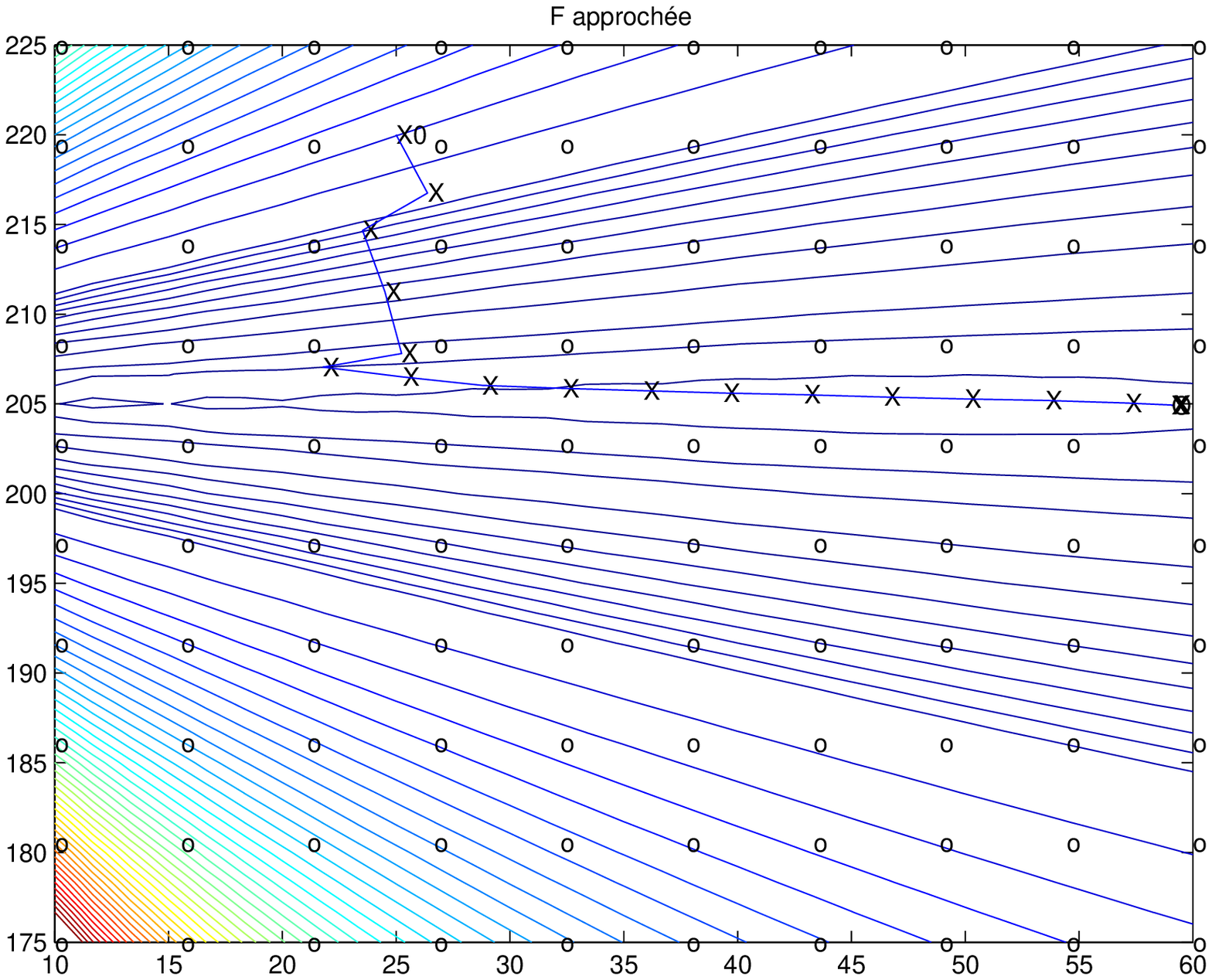} \\
Grid ($5 \times 5$) & Grid ($10 \times 10$) \\
\vspace{1mm}
Solution : \quad \parbox[t]{3cm}{$F = 60.000$ \\ $M = 205.26$} &
Solution : \quad \parbox[t]{3cm}{$F = 59.073$ \\ $M = 204.91$} \\
38 evaluations of AD & 38 evaluations of AD \\
\\ \\
\vspace{3mm}
\includegraphics*[width=6.5cm]{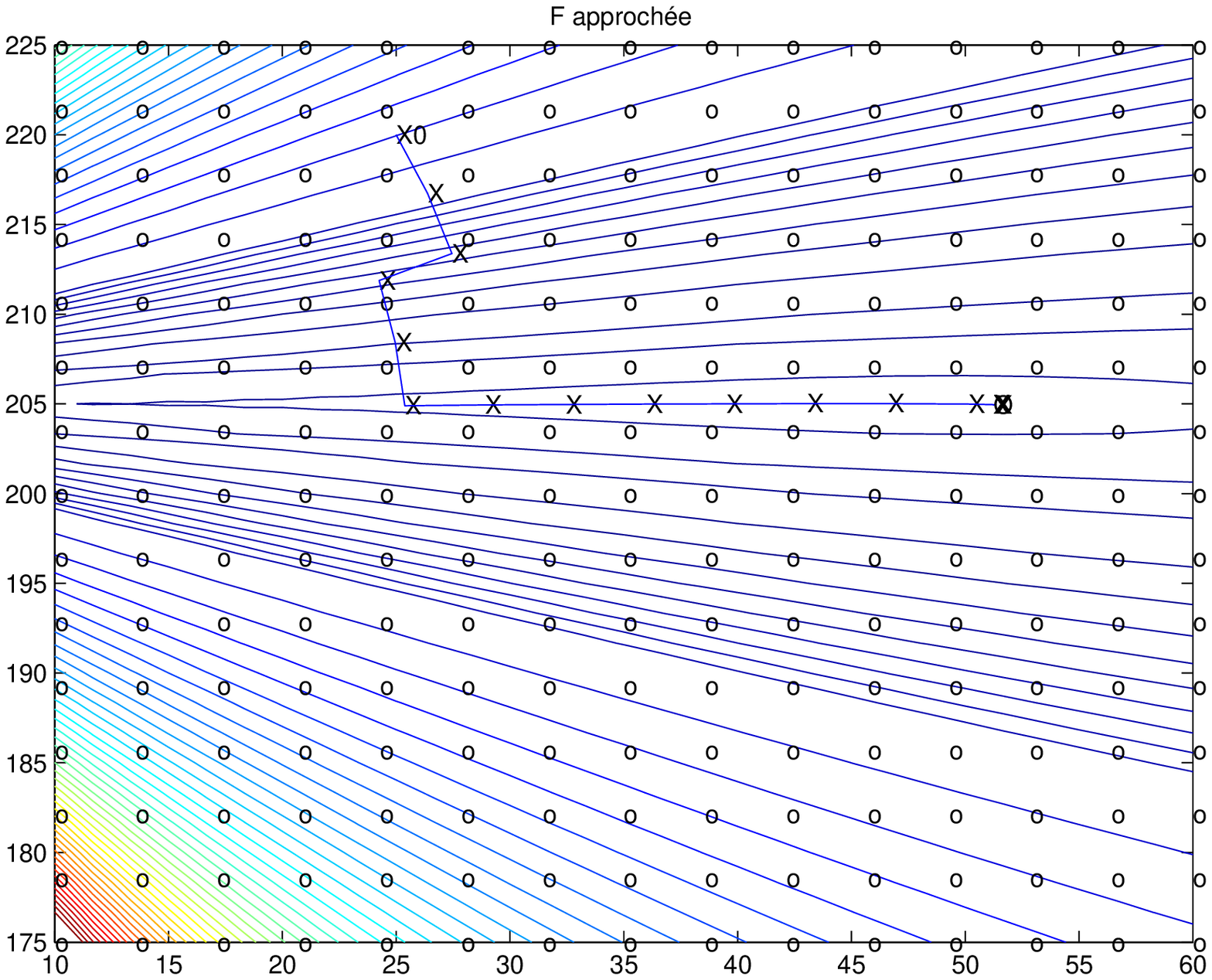} &
\includegraphics*[width=6.5cm]{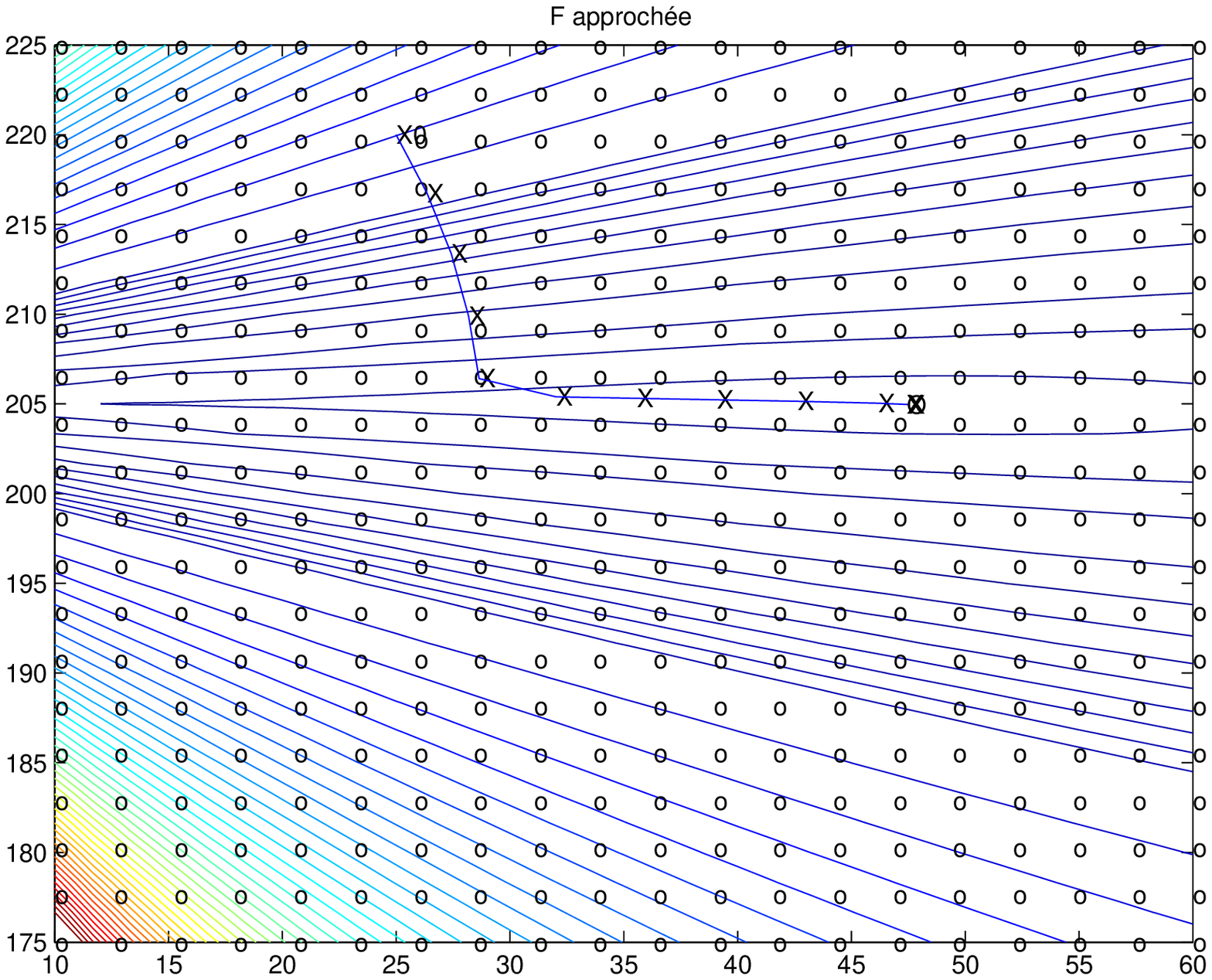} \\
Grid ($15 \times 15$) & Grid ($20 \times 20$) \\
\vspace{1mm}
Solution : \quad \parbox[t]{3cm}{$F = 51.218$ \\ $M = 204.95$} &
Solution : \quad \parbox[t]{3cm}{$F = 47.444$ \\ $M = 204.97$} \\
19 evaluations of AD & 15 evaluations of AD \\
\end{tabular}
\caption{T letter cantilever: Gradient method iterative
computation on a grid.} \label{fig_AD}
\end{figure*}

The first solution is obtained by the diffuse approximation based
gradient method. The calculation of the cost function is first
carried out for all the 'nodes' of the following grids: $5 \times
5; 10 \times 10; 15 \times 15\ \mbox{and}\ 20 \times 20$. The
gradient type procedure is then started on the grid and, thanks to
the smoothness of the diffuse approximation base representation of
the approximate value of the cost function, converged with no
difficulty in roughly 20-40 iterations. The iterative values
obtained in the gradient method computations are for different
grids shown in Figure \ref{fig_AD}. Grid is constructed for the
following interval of values of force and moment: $$F \in \langle
10, 60\rangle\ ; \quad M \in \langle175, 225\rangle.$$

We note that all different choices of the grid can result in different
solutions, since none of the solutions of this kind will satisfy the
equilibrium equations.

\begin{figure*}[hbt]
\centering
\begin{tabular}{c@{\hspace{3mm}}c}
\fbox{\includegraphics*[width=6.5cm]{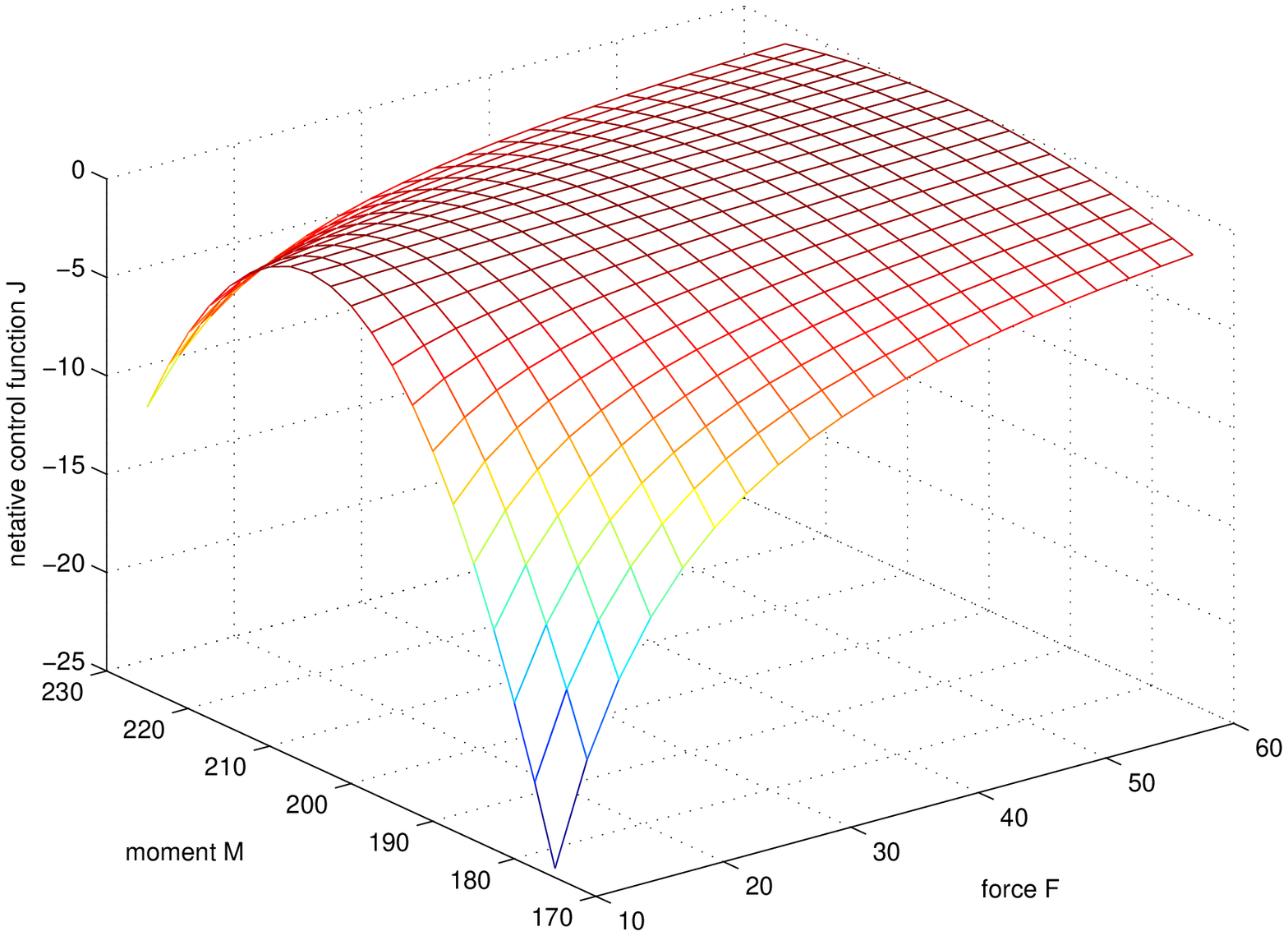}} &
\fbox{\includegraphics*[width=6.5cm]{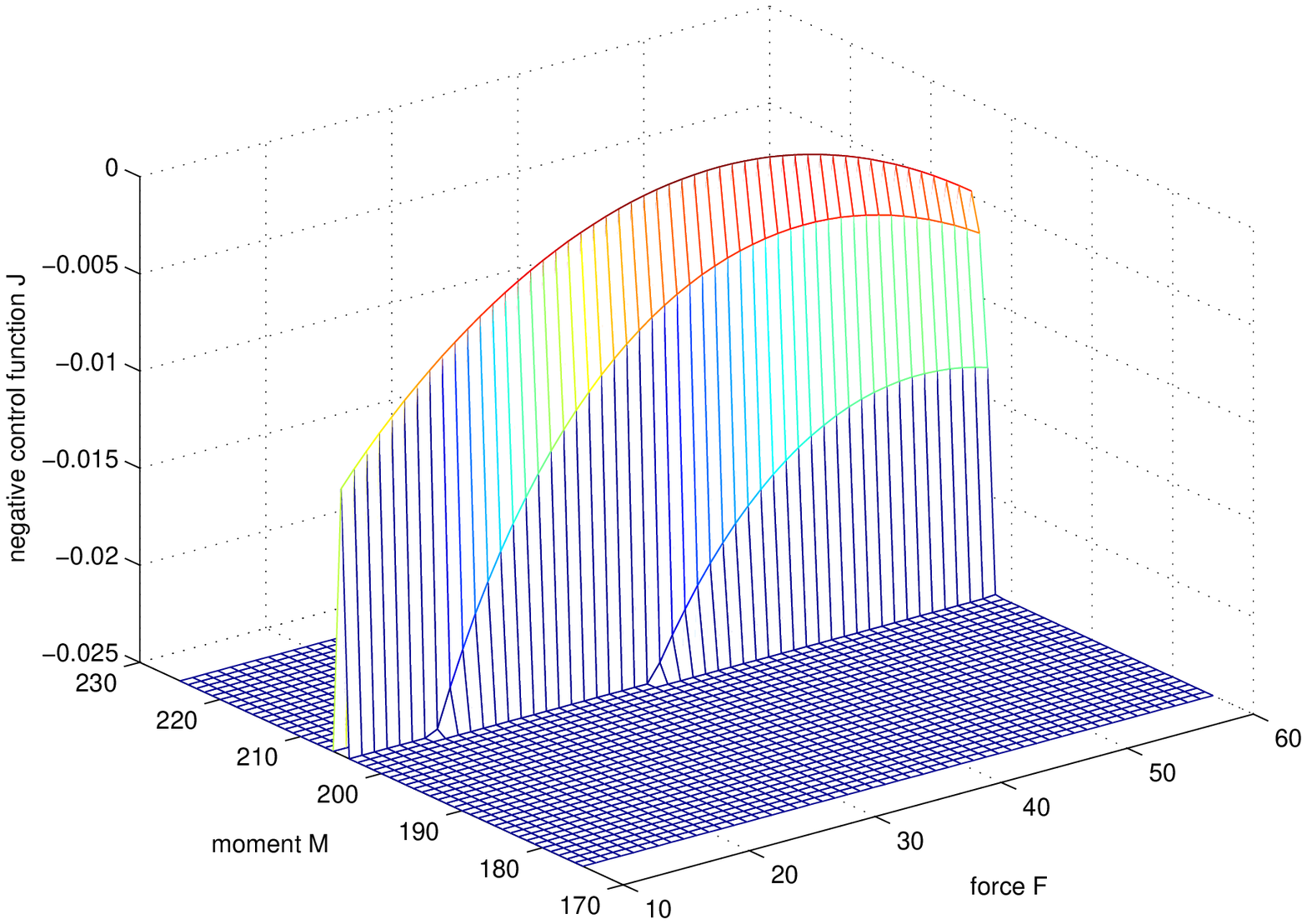}} \\
\\
Whole scale of values. & More detailed about value of optimum. \\
\end{tabular}
\caption{T letter cantilever: contour of the cost function.}
\label{tecko2}
\end{figure*}

Quite large difference between known optimal solution and solutions of response
surface could be explained by different influence of each control variable to
value of cost function, see Figure \ref{tecko2}. All used grids weren't able to
describe small influence of force $F$.

The second solution method used for this problem employs the
genetic algorithm based computation. In this computation we have
used the same admissible intervals like in the previous case for
the control variables, force and moment. The computations are
carried out starting from the random values chosen in this
interval and stopped when the first value of the cost function $J
\leq 10^{-7}$ is found.

In order to be able to look into statistics, one hundred runs are
performed with each one converging to the exact solution. Three
types of procedures were tried, with either tuning on or off the
parameters controlling local mutation (LM) and the gradient-type
acceleration of sign change (SG). The results are presented in
Table \ref{rslt-T-sade}.

\begin{table}[h]
\begin{center}
\begin{tabular}{|l|rrr|}
\hline
\multicolumn{4}{|c|}{\bf Letter T problem} \\
\hline
Computation & \multicolumn{3}{|c|}{Number of fitness calls} \\
\cline{2-4}
with SADE & Minimum & Maximum & \bf Mean Value \\
\hline
LM-on SG-off & 240 & 2860 & \bf 648.8\\
LM-off SG-off & 280 & 1680 & \bf 625.4  \\
LM-off SG-on & 220 & 1640 & \bf 591.2  \\
\hline
\end{tabular}
\end{center}
\caption{T-letter cantilever: SADE algorithm performance}
\label{rslt-T-sade}
\end{table}

We can see that the best performance is achieved with the simple gradient-like
modification of SADE genetic algorithm to accelerate convergence in the latest
stage. The results obtained with this simple gradient-like modification of SADE
algorithm can further be improved by using described GRADE algorithm, with a
special role of the radioactivity parameter. See Table \ref{rslt-T-grade}.

\begin{table}[h]
\begin{center}
\begin{tabular}{|l|rrr|}
\hline
\multicolumn{4}{|c|}{\bf Letter T problem} \\
\hline
Computation & \multicolumn{3}{|c|}{Number of fitness calls} \\
\cline{2-4}
with GRADE & Minimum & Maximum & \bf Mean Value \\
\hline
radioactivity = 0.0 & 180 & 880 & \bf 410.0 \\
radioactivity = 0.1 & 240 & 1200 & \bf 440.8  \\
radioactivity = 0.2 & 280 & 1180 & \bf 512.4 \\
radioactivity = 0.3 & 280 & 1300 & \bf 588.4 \\
\hline
\end{tabular}
\end{center}
\caption{T letter cantilever: GRADE algorithm performance}
\label{rslt-T-grade}
\end{table}

The second solution procedure applied to the same example is
so-called simultaneous solution of mechanics equilibrium and
optimal control equations, written explicitly in (\ref{eq_4.14})
to (\ref{eq_4.16}) with value $\alpha = 0$. The total numb er of
unknowns in this case is equal to 44, with 2 control variables
(the force and the moment), 21 components of nodal displacements
and rotations and 21 Lagrange multiplier. For computational
convenience, the problem of solving the set of nonlinear algebraic
equations in (\ref{eq_4.14}) to (\ref{eq_4.16}) is recast as the
minimization statement which allows direct application of the
genetic algorithm with
\begin{equation}
{\bf r} := \left[
\begin{array}{c} {\bf r}^{\lambda} \\ {\bf r}^{\phi} \\ {\bf r}^{\nu}
\end{array}
\right] = {\bf 0} \Leftrightarrow \underset{({\bf \varphi}, {\bf \nu},
{\bf \lambda})}{\min} {\bf r}^T{\bf r}
\end{equation}

The solution efficiency of the proposed simultaneous procedure
depends on the chosen upper and lower bounds of the admissible
interval and the initial guess for the solution. For example, the
mechanics state variables are chosen as these featuring in the
desired beam shape, $\varphi^d$, and the bounds are controlled by
the chosen parameter $EP$ according to
\begin{equation}
\bmvarphi \in \left[ (1-EP)\bmvarphi^d, (1+EP)\bmvarphi^d \right]
\end{equation}

The equivalent bounds on control variables are obtained from
the precious result obtained by solving the grid minimization problem which
results with the minimum of the response surface. Finally, the Lagrange
multipliers is solved for from (\ref{eq_4.15}) where the adopted values for
$\bmvarphi$ and $\bmnu$ are chosen.

One hundred computations is performed with the indicated bounds on
the unknowns. Value of $EP$ parameter is set to $0.0001$. Choice
of parameters of GRADE algorithm were: $PR = 30, CL = 2$ and
'radioactivity' equal to $0.2$. Table \ref{tab_sim} summarizes the
statistics of this computation.
\begin{table}[hbt] \centering
\begin{tabular}{c|rrrl}
 & Minimum & Maximum & Mean Value & Standard deviation\\
\hline
 $F$ & 39.957 & 40.048 & 40.000 & 0.0199 \\
 $M$ & 204.86 & 205.15 & 205.00 & 0.0548 \\
\hline
number of evaluations of $J(\cdot)$ & 100740 & 298080 & 172176 & ---------
\end{tabular}
\caption{T letter cantilever : solution statistics}
\label{tab_sim}
\end{table}

The last part of the analysis carried out in this examples
concerns an attempt to further increase the efficiency of the
simultaneous solution procedure. In that sense, we employ the
GRADE version of genetic algorithm with the choice of parameters
$PR = 20, CL = 1$ and 'radioactivity' equal to $0.2$. A very small
value of bounds is chosen as well with $EP = 0.00001$. The
computations we have performed with a hundred runs of the genetic
algorithm can be summarized with the statistics given in Table 4.

\begin{table}[hbt]
\centering
\begin{tabular}{c|rrrl}
 & Minimal & Maximal & Mean Value & Standard deviation\\
\hline
 $F$ & 39.973 & 40.034 & 40.000 & 0.0135 \\
 $M$ & 204.96 & 205.05 & 205.00 & 0.0192 \\
\hline
number of evaluations of $J(\cdot)$ & 14720 & 201480 & 37701 & ---------
\end{tabular}
\caption{T Letter cantilever : solution statistics}
\label{tab_sim2}
\end{table}

We can see from Table 4 that the proper choice of the bounds can force the
algorithm to always converge to the same solution. The latter is the consequence
of using the simultaneous solution procedure which assures that the computed
solution also satisfies the equilibrium equations. Moreover, the total cost of
the simultaneous solution procedure can be reduced beyond the one needed for
response-surface-based approximate solution computations, by either reducing the
interval as done herein or by making gradient-type modification of this
algorithm in order to accelerate the convergence rate.

\subsection{Optimal control of a cantilever structure in form of letter I}

\begin{figure}[!ht]
\vspace{9mm}
\begin{center}
\includegraphics[width=100mm,keepaspectratio]{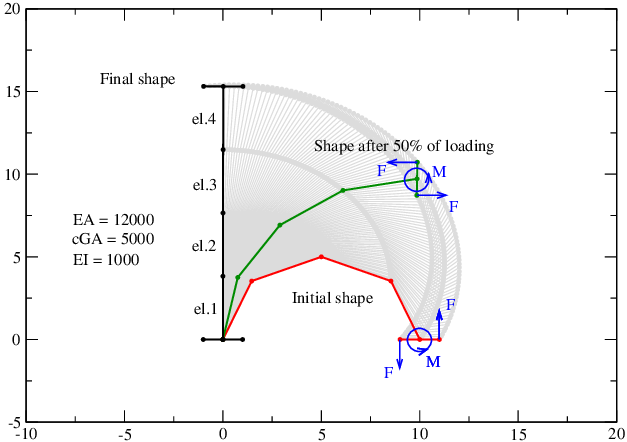}
\caption{I letter cantilever: initial, final and intermediate configurations}
\label{icko1}
\end{center}
\end{figure}

In the second example we deal with a problem which has a multiple
solution and its regularized form which should restore the
solution uniqueness. To that end, a cantilever beam is used very
much similar to the one studied in the previous example, except
for a shorter straight bar with length equal $2$. The cantilever
is controlled by a moment $M$ and a couple of follower forces
which follow the rotation of the cross-sections to which they are
attached. The initial and final configuration, which is obtained
for a zero couple and a moment $M=205.4$ are shown in Figure
\ref{icko1}, along with a number of intermediate configurations.

The first computation is performed with the cost function
identical to the one in (\ref{eq_contr_discr}), imposing only the
minimum of difference between the desired and the computed
deformed shape, with no restriction on control variables. The
computation is carried out by using the GRADE genetic algorithm by
starting with random values within the selected admissible
intervals for the force couple and moment according to
$$F \in \langle 0, 20 \rangle\ ; \quad M \in \langle 0, 230 \rangle $$
The algorithm performance is illustrated in Table
\ref{rslt-I-grade}.
\begin{table}[htb]
\begin{center}
\begin{tabular}{|rrr|}
\hline
\multicolumn{3}{|c|}{\bf Letter I problem} \\
\hline
\multicolumn{3}{|c|}{Number of fitness calls} \\
\hline
Minimum & Maximum & \bf Mean Value \\
\hline
180 & 640 & \bf 359.8 \\
\hline
\end{tabular}
\end{center}
\caption{I letter cantilever: GRADE algorithm performance}
\label{rslt-I-grade}
\end{table}

\begin{figure}[htb]
\vspace{10mm}
\begin{center}
\includegraphics[width=100mm,keepaspectratio]{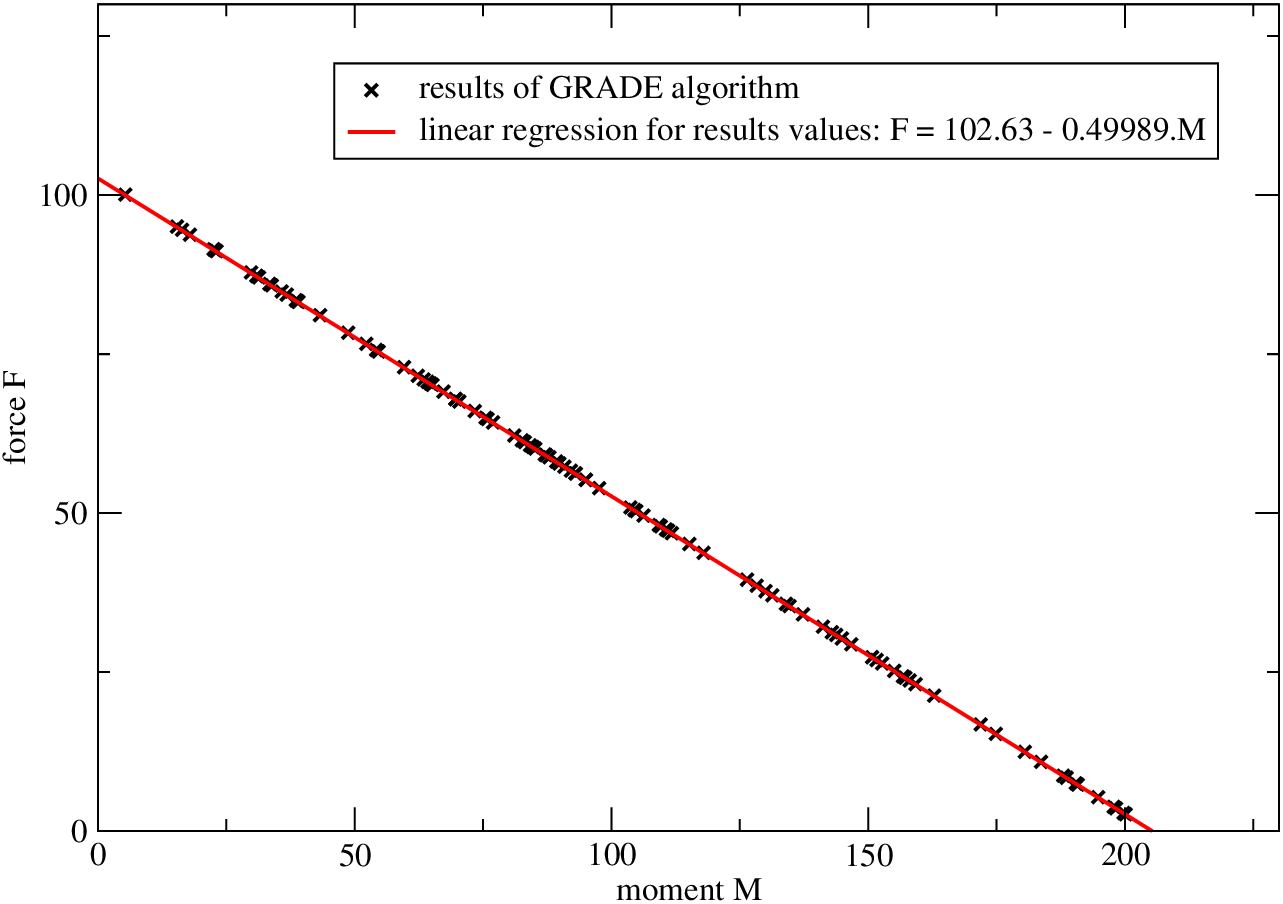}
\caption{I letter cantilever: 100 different solutions}
\label{fig_I_inf}
\end{center}
\end{figure}

A number of different solutions have been obtained with different
computer runs which were performed (see Figure \ref{fig_I_inf}).
However, all these solutions remain in fact clearly related
considering that the applied moment and the force couple play an
equivalent role in controlling the final deformed shape; It can be
shown for this particular problem that any values of force and
moment which satisfy a slightly perturbed version (because of the
straight bar flexibility) of the force equilibrium
$$F \cdot h + M = \bar{M} = 205.4$$
will be admissible solution, thus we have infinitely many
solutions for the case where only the final shape is controlled by
the choice of the cost function.

In order to eliminate this kind of problem we can perform a regularization of
the cost function, by requiring not only that the difference between computed
and final shape be minimized but also that control variables be as small as
possible; Namely, with a modified form of the cost function
\begin{equation}
J^h(\hat{\bf u}_a(\bmnu),\bmnu) =  \frac{1}{4} \sum_{e=0}^{n_{el}} \sum_{a=1}^2
l^e \left( \hat{\bf u}_a(\bmnu) - {\bf u}_a^d \right)^T \left( \hat{\bf
u}_a(\bmnu) - {\bf u}_a^d \right) + \alpha \bmnu^T \bmnu
\label{eq_contr_I2}
\end{equation}
where $\alpha$ is a chosen weighting parameter specifying the
contribution of the control. We set a very small value $\alpha =
10^{-9}$ and choose the convergence tolerance to $10^{-12}$ and
carry out the computation for yet a hundred times. Whereas a more
stringent value of the tolerance requires somewhat larger number
of cost function evaluation, the result in each of the runs
remains always the same, given as
$$ F = 68.466\ ; \quad M = 68.526 $$
and the found optimal value of the cost function is $J = 1.4078631.10^{-5}$.

\begin{table}[htb]
\begin{center}
\begin{tabular}{|rrr|}
\hline
\multicolumn{3}{|c|}{\bf Letter I problem extended} \\
\hline
\multicolumn{3}{|c|}{Number of fitness calls} \\
\hline
Minimum & Maximum & \bf Mean Value \\
\hline
820 & 16320 & \bf 1758.8 \\
\hline
\end{tabular}
\end{center}
\caption{I letter cantilever: GRADE algorithm performance}
\label{rslt-icko2-grade}
\end{table}

\subsection{Optimal control of deployment of a multibody system}

\begin{figure}[hbt]
\vspace{8mm}
\begin{center}
\includegraphics[width=100mm,keepaspectratio]{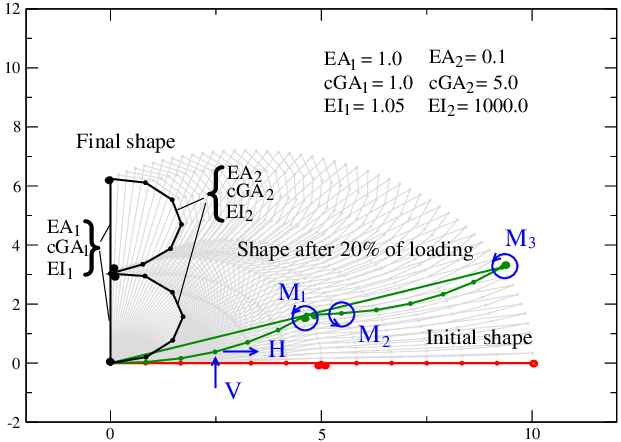}
\caption{Multibody system deployment: initial, final and intermediate
configurations.}
\label{becko1}
\end{center}
\end{figure}

The optimal control procedure of the deployment problem of a
multibody system is studied in this example. In its initial
configuration the multibody system consists of two flexible
component (with $EA = 0.1$, $GA = 0.05$ and $EI = 10^3$) each 5
units long connected by the revolute joints (see
\cite{Ibrahimbegovic:Mamouri:2000}) to a single stiff component
(with $EA = 1$, $GA = 0.5$ and $EI = 10^5$) of the length equal to
$10$, which are all placed parallel to horizontal axis. In the
final deployed configuration the multibody system should take the
form of a letter B with the stiff component being completely
vertical and two flexible component considerably bent. The
deployment of the system is controlled by five control variables,
three moments $M_1$, $M_2$ and $M_3$, a vertical $V$ and a
horizontal force $H$. See Figure \ref{becko1}.

The cost function in this problem is again chosen as the one in
\ref{eq_contr_discr}, which controls that the system would find the
configuration as close as possible to the desired configuration. The desired
configuration of the system corresponds to the values of forces $H = 0.04$, $V =
-0.05$ and moments $M_1 = 0.782$, $M_2 = -0.792$ and $M_3 = 0.792$. The solution
is computed by using the SADE and GRADE genetics algorithms and starting with
the random choice in the interval of interest defined as
$$H \in \langle 0.025; 0.05 \rangle , \quad V \in \langle -0.06; -0.035 \rangle
, \quad M_1 \in \langle 0.6; 0.9 \rangle , $$
$$\quad M_2 \in \langle -0.9; -0.65
\rangle , \quad M_3 \in \langle 0.6; 0.85 \rangle .$$

\begin{figure}[!ht]
\centering
\begin{tabular}{c@{\hspace{3mm}}c}
\fbox{\includegraphics*[width=6.5cm]{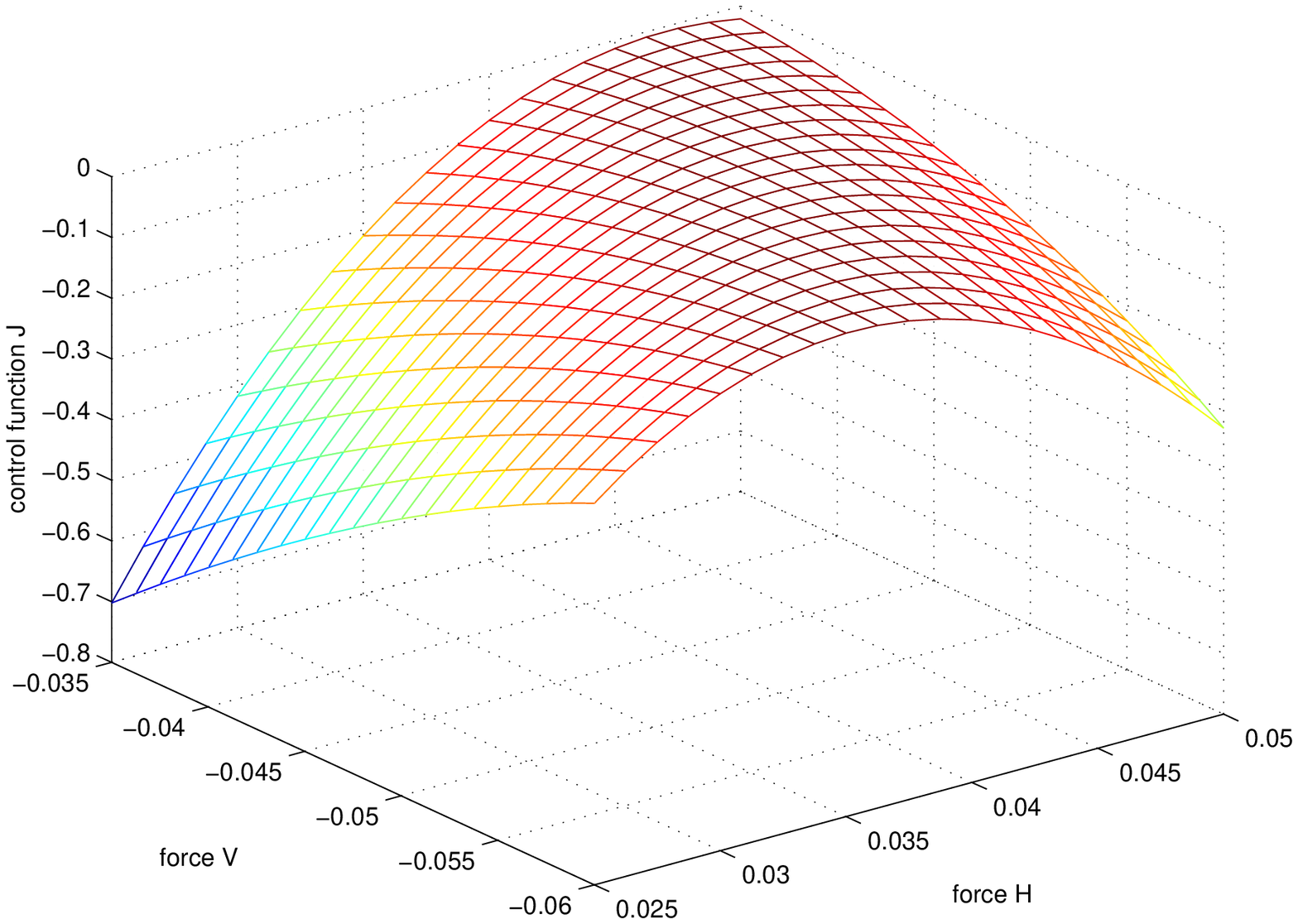}} &
\fbox{\includegraphics*[width=6.5cm]{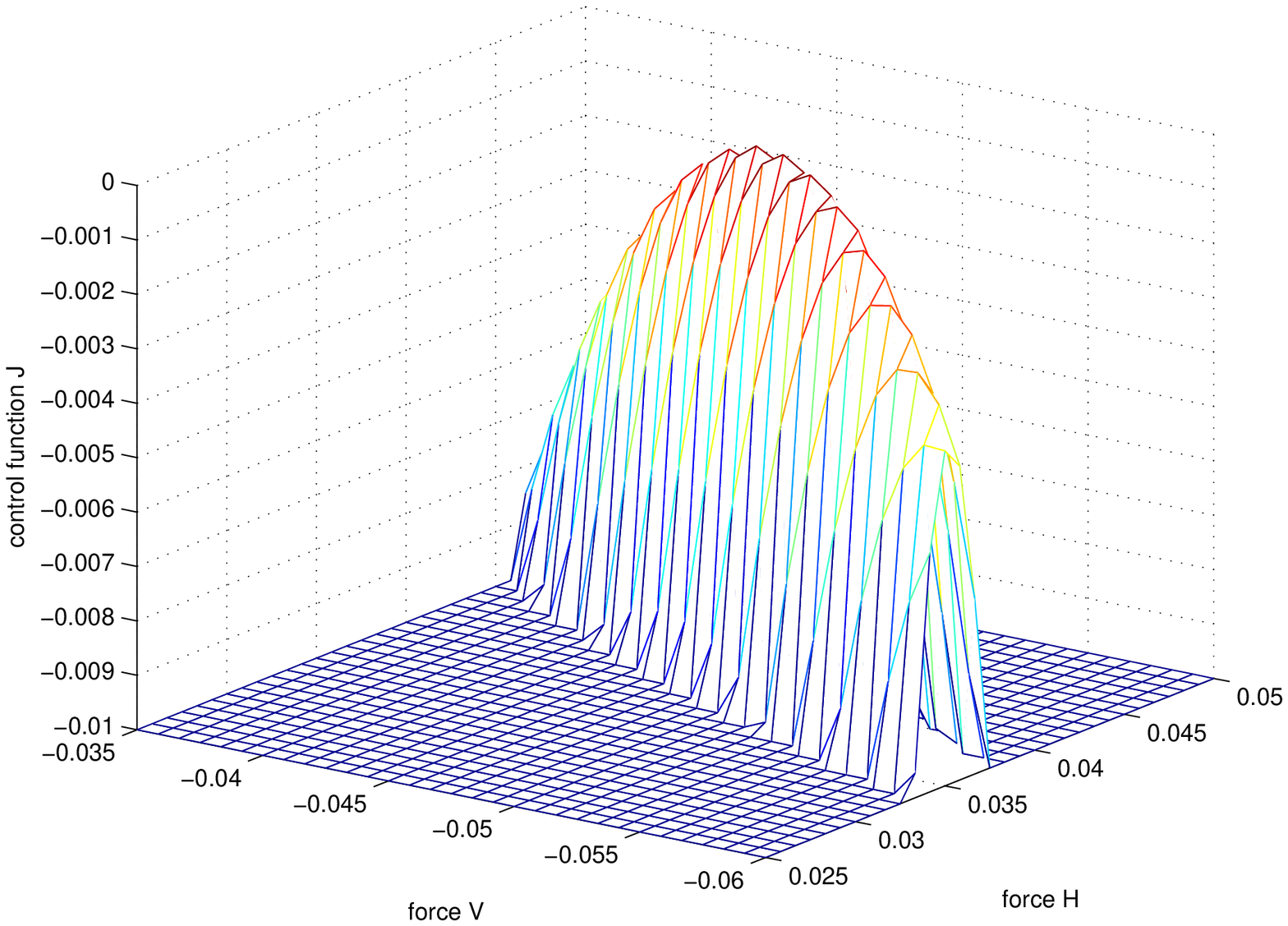}} \\
\\
$H-V$ subspace & $H-V$ subspace  \\
\\
\fbox{\includegraphics*[width=6.5cm]{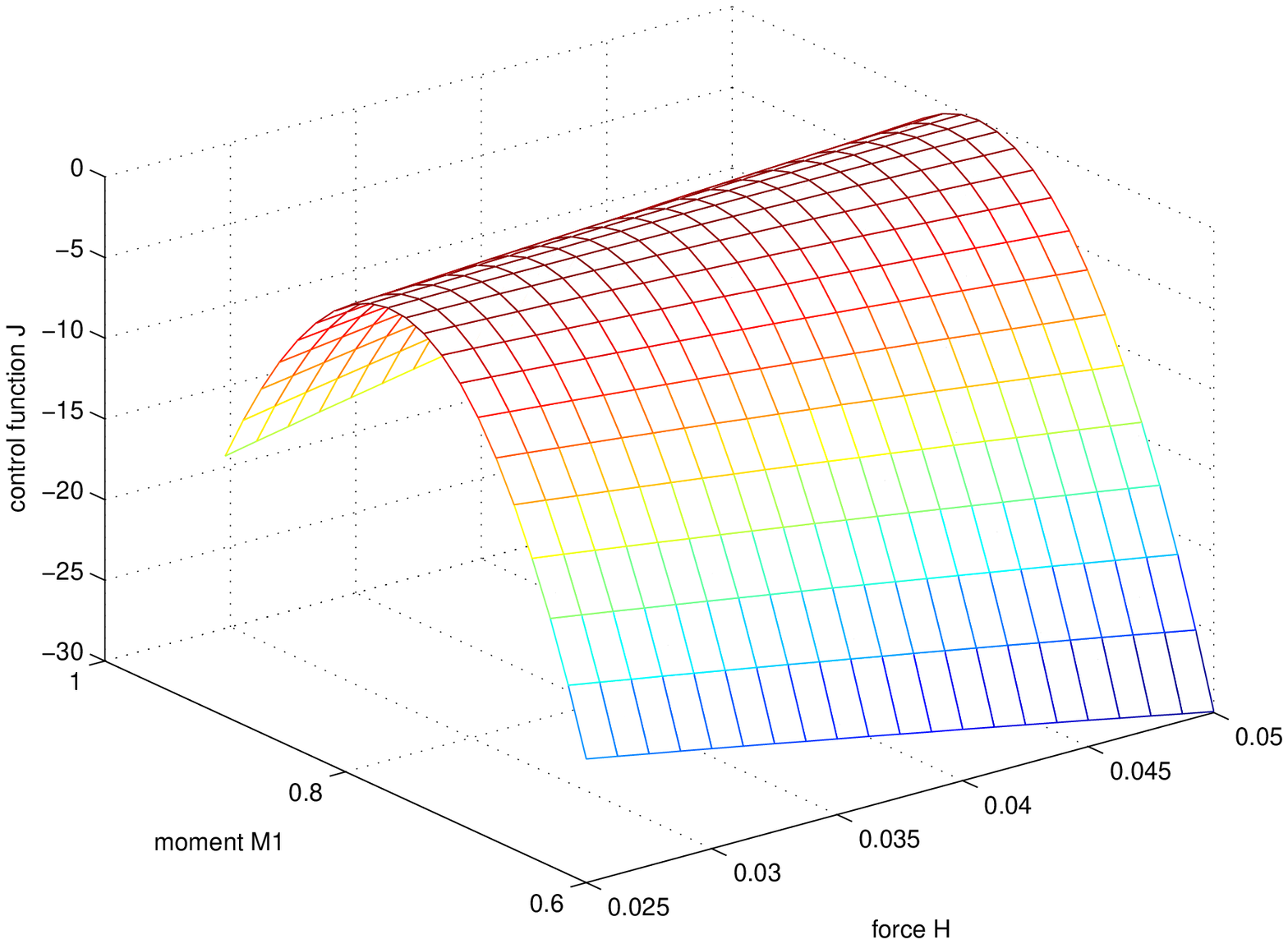}} &
\fbox{\includegraphics*[width=6.5cm]{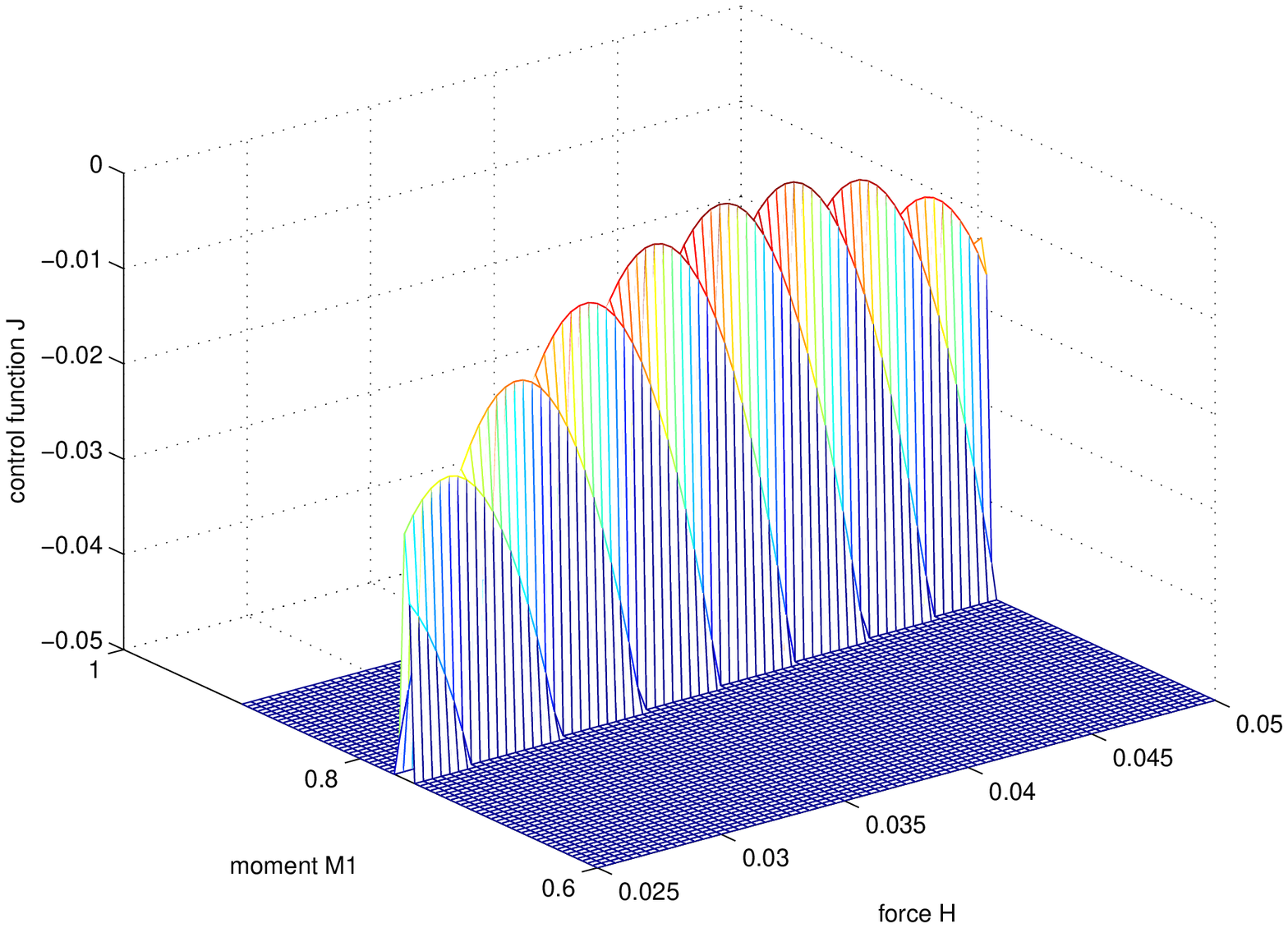}} \\
\\
$H-M_1$ subspace & $H-M_1$ subspace  \\
\\
\fbox{\includegraphics*[width=6.5cm]{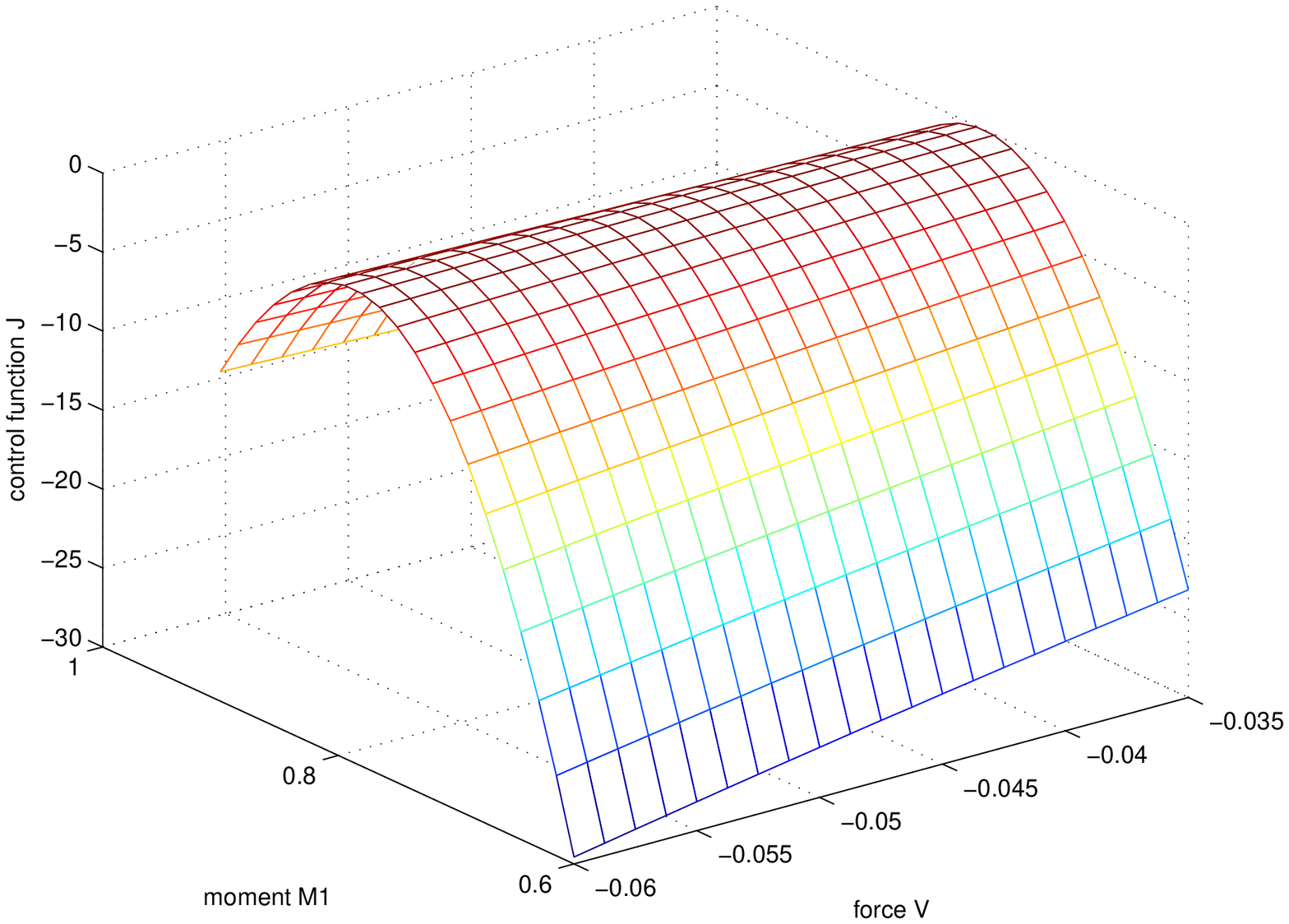}} &
\fbox{\includegraphics*[width=6.5cm]{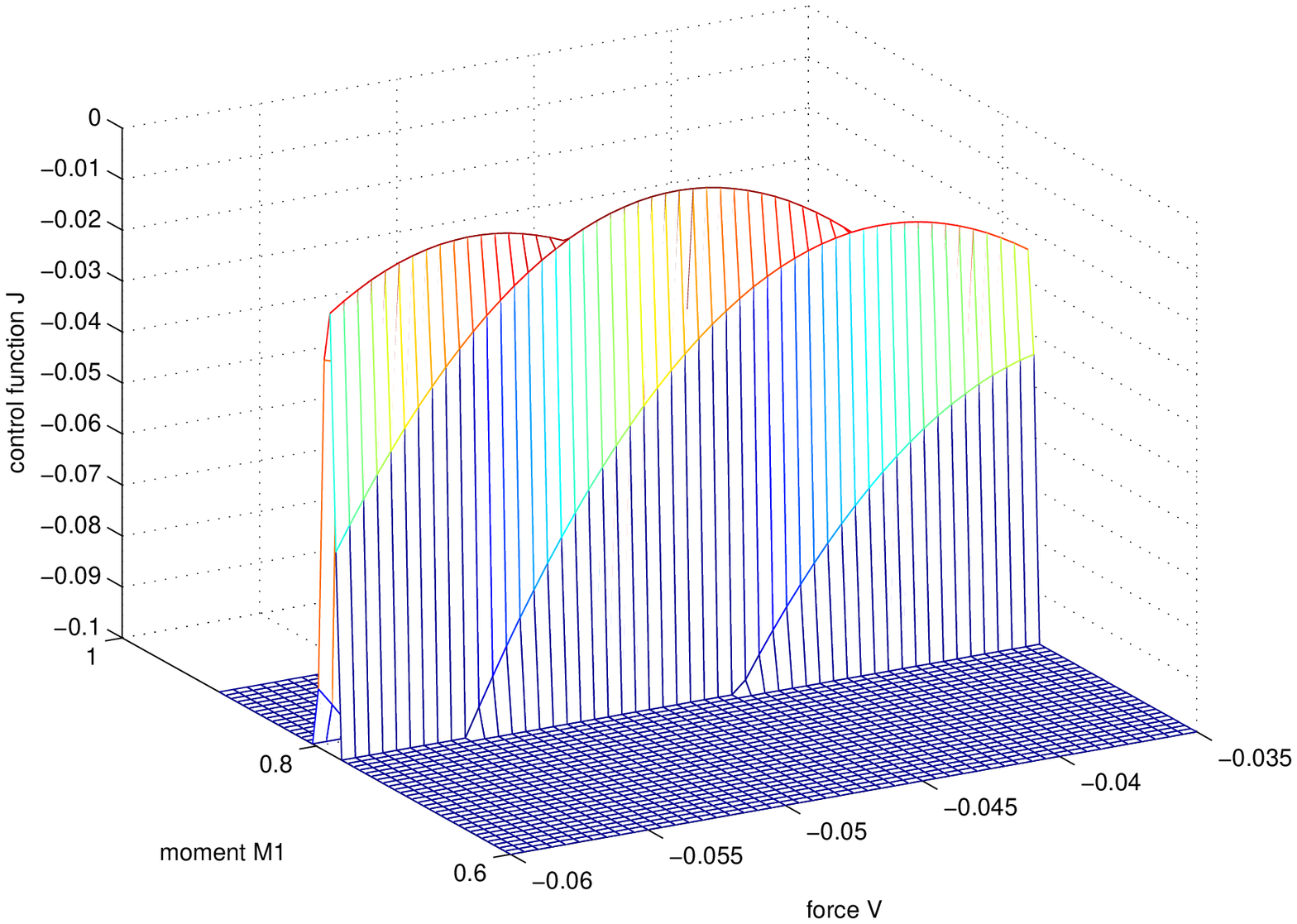}} \\
\\
$V-M_1$ subspace & $V-M_1$ subspace
\end{tabular}
\caption{Multibody system deployment: contours of the cost
function in different subspaces.} \label{becko2a}
\end{figure}

\begin{figure}[!ht]
\centering
\begin{tabular}{c@{\hspace{3mm}}c}
\fbox{\includegraphics*[width=6.5cm]{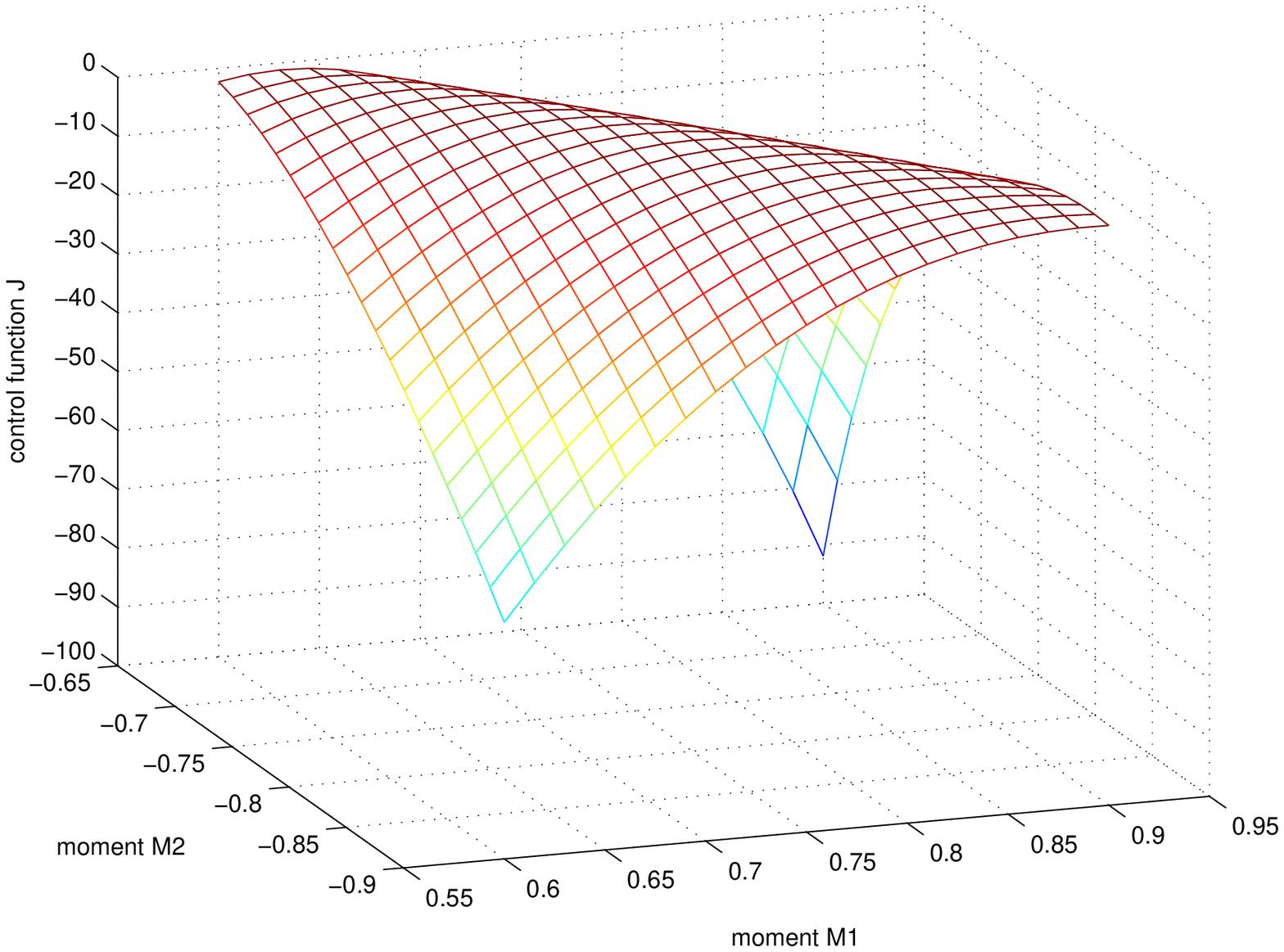}} &
\fbox{\includegraphics*[width=6.5cm]{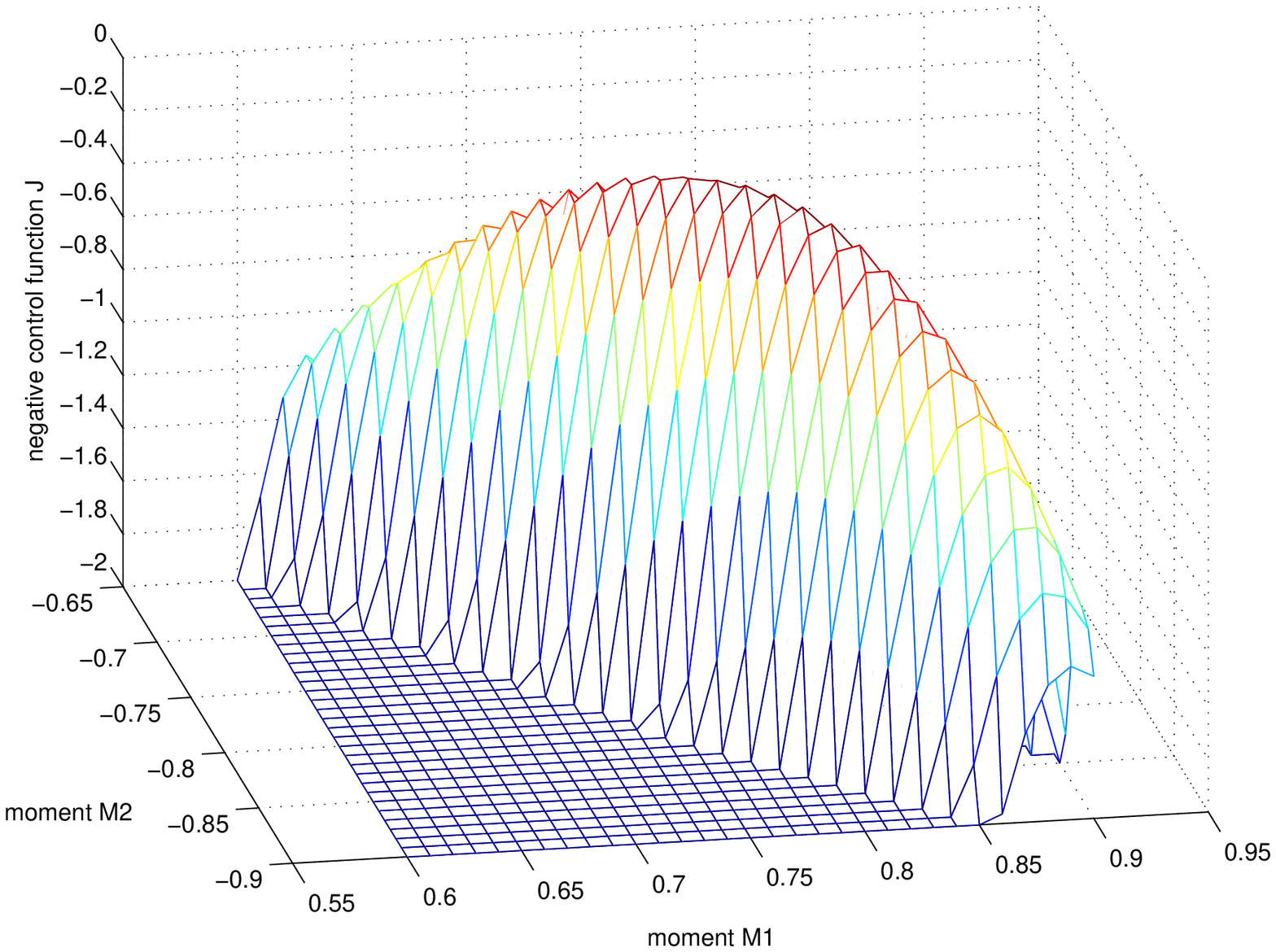}} \\
\\
$M_1-M_2$ subspace & $M_1-M_2$  subspace \\
\\
\fbox{\includegraphics*[width=6.5cm]{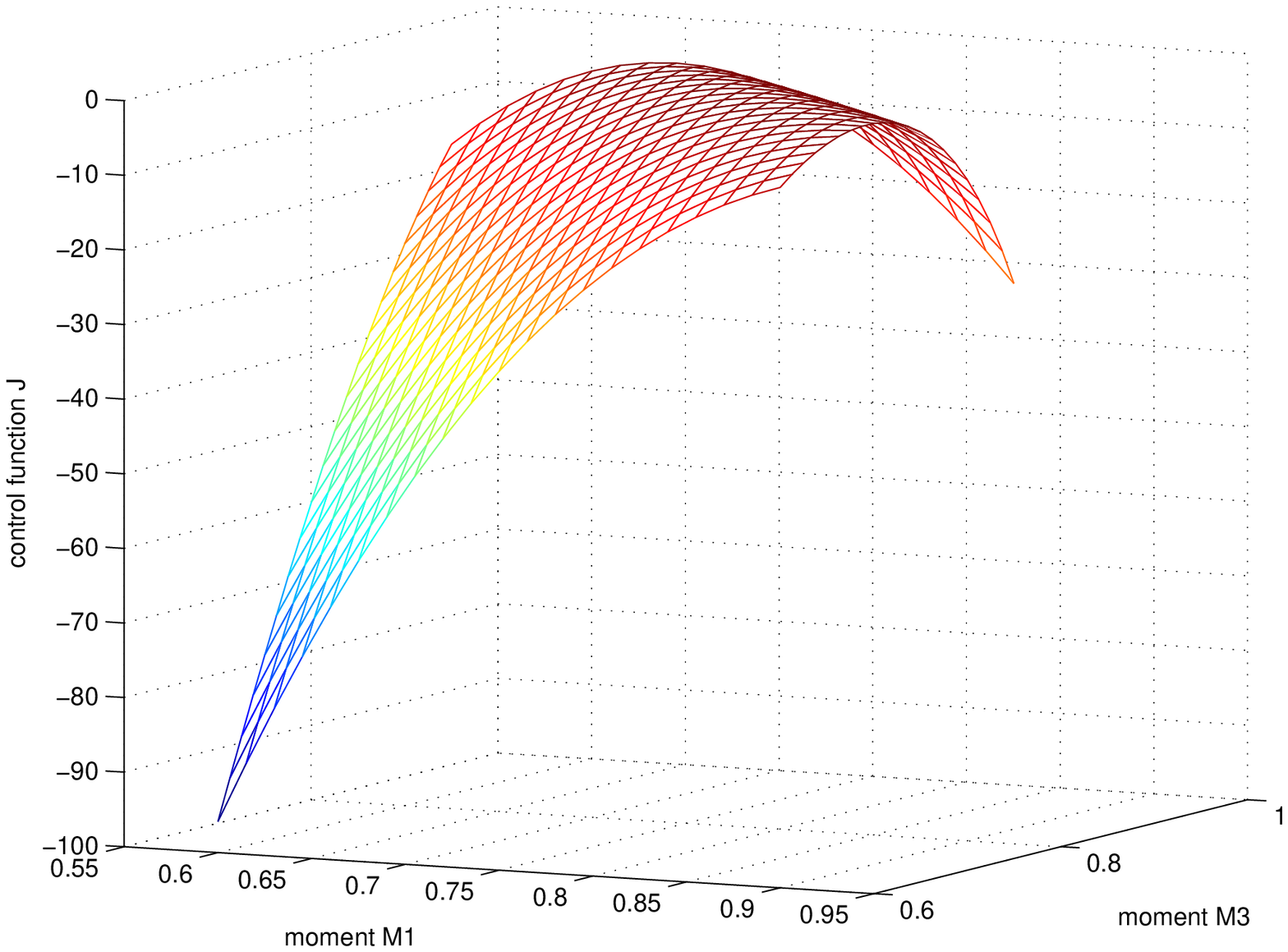}} &
\fbox{\includegraphics*[width=6.5cm]{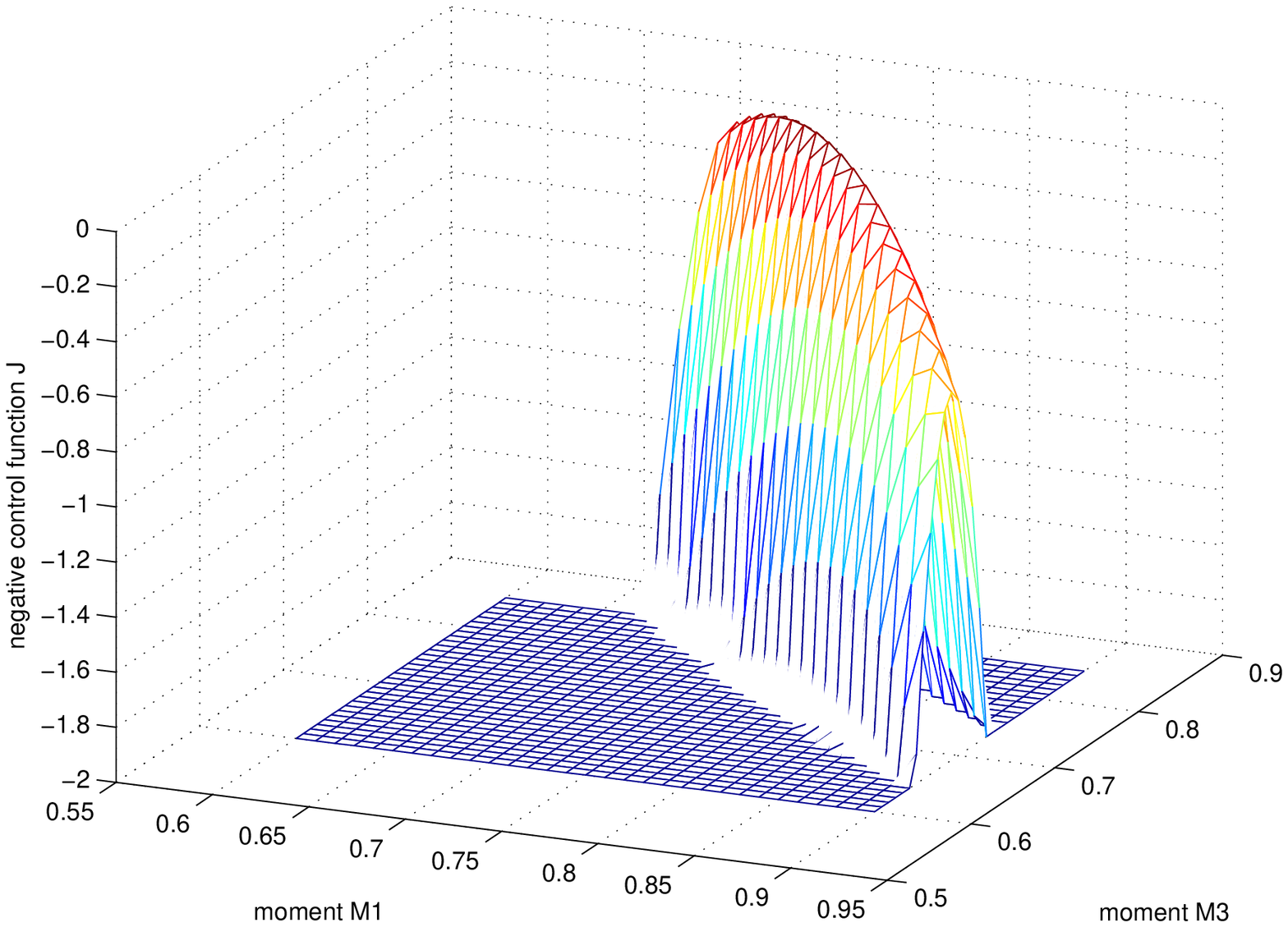}} \\
\\
$M_1-M_3$  subspace & $M_1-M_3$  subspace \\
\\
\fbox{\includegraphics*[width=6.5cm]{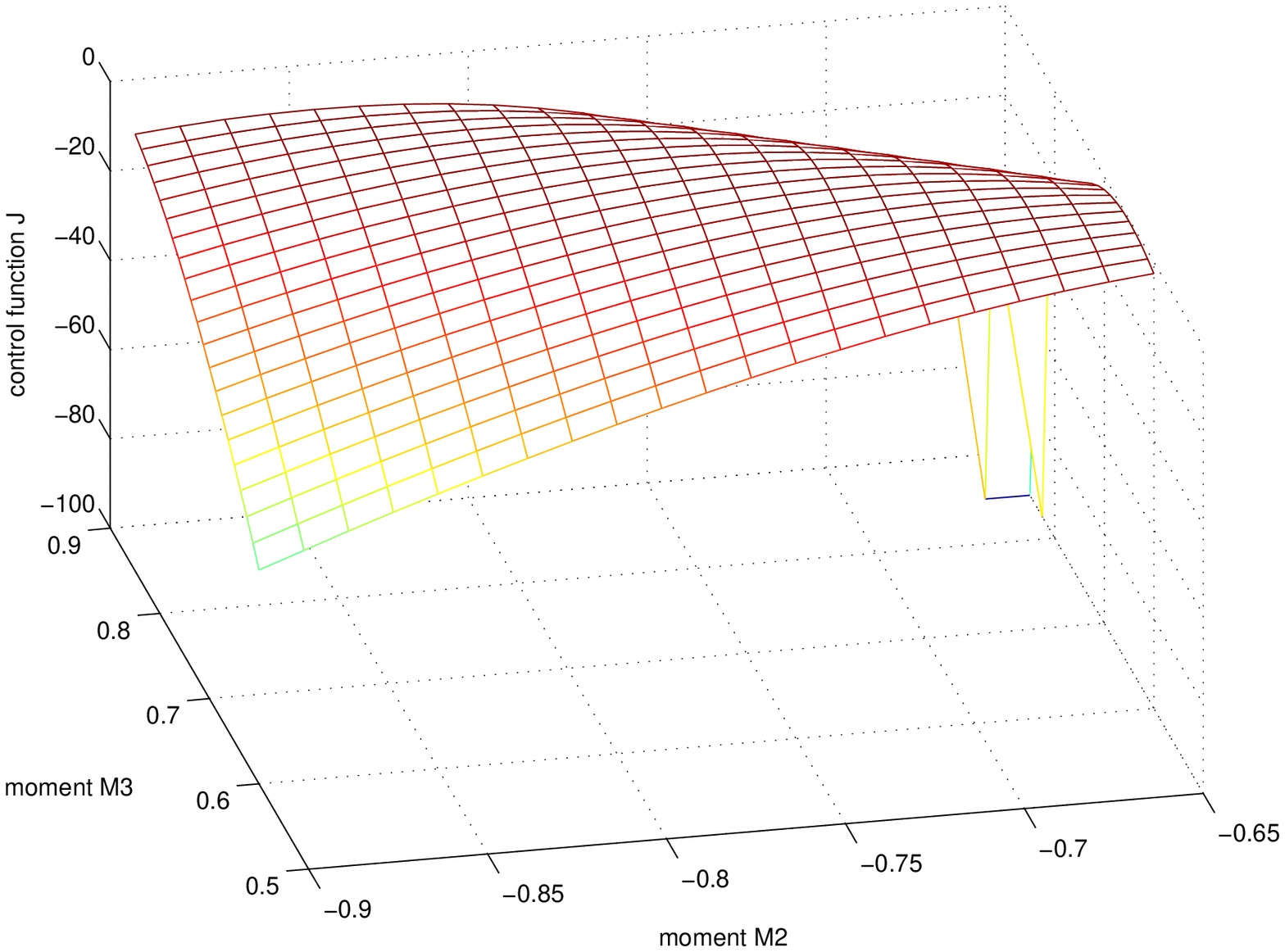}} &
\fbox{\includegraphics*[width=6.5cm]{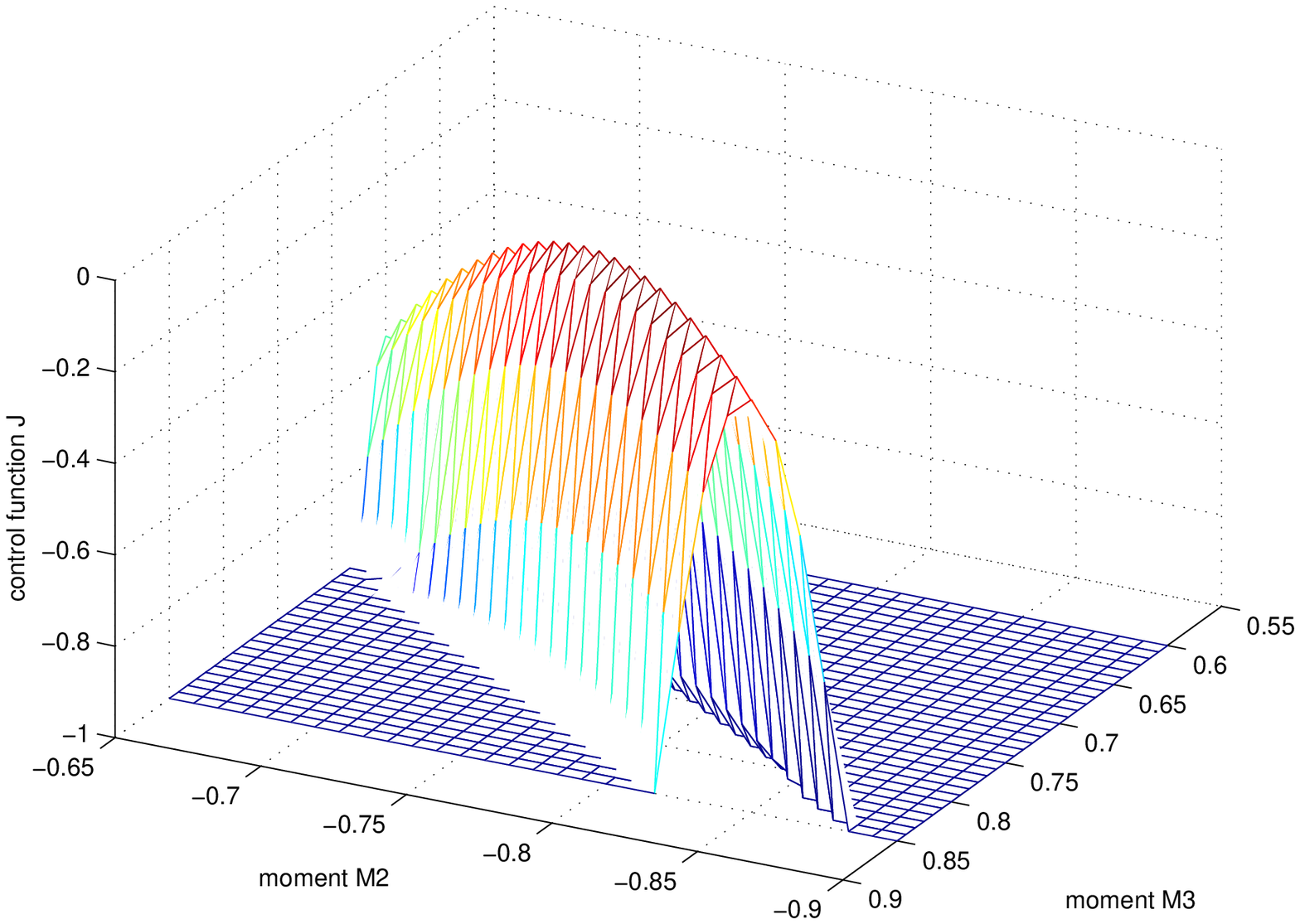}} \\
\\
$M_2-M_3$  subspace & $M_2-M_3$  subspace
\end{tabular}
\caption{Multibody system deployment: contours of the cost
function in different subspaces.} \label{becko2b}
\end{figure}

The solution of the problem is typically more difficult to obtain with an
increase in the number of control variables, one of the reasons being a more
irregular form of the cost function. In that sense,we refer to illustrative
representation of the cost function contours in different subspaces of control
variables as shown in Figures \ref{becko2a} and \ref{becko2b}.

The convergence tolerance on cost function is chosen equal to $10^{-6}$. The
SADE algorithm performance for the simplest choice of algorithm parameters is
presented in Table \ref{rslt-B-sade} and the GRADE algorithm
performance for modified value of radioactivity parameter is
presented in Table \ref{rslt-B-grade}.

\begin{table}[h]
\begin{center}
\begin{tabular}{|l|rrr|}
\hline
\multicolumn{4}{|c|}{\bf Letter B problem} \\
\hline
Computation & \multicolumn{3}{|c|}{Number of fitness calls} \\
\cline{2-4}
with SADE & Minimum & Maximum & \bf Mean Value \\
\hline
LM-on SG-off & 2600 & 165800 & \bf 46887.5 \\
LM-off SG-off & 2350 & 177150 & \bf 42494.0  \\
LM-off SG-on & 2400 & 177850 & \bf 34612.1  \\
\hline
\end{tabular}
\end{center}
\caption{Results of SADE algorithm for 5-variable control problem}
\label{rslt-B-sade}
\end{table}

\begin{table}[h]
\begin{center}
\begin{tabular}{|l|rrr|}
\hline
\multicolumn{4}{|c|}{\bf Letter B problem} \\
\hline
Computation & \multicolumn{3}{|c|}{Number of fitness calls} \\
\cline{2-4}
with GRADE & Minimum & Maximum & \bf Mean Value \\
\hline
radioactivity = 0.0 & ----- & ----- & \bf ----- \\
radioactivity = 0.1 & 1500 & 114050 & \bf 23862.5  \\
radioactivity = 0.2 & 1900 & 117850 & \bf 20632.0  \\
radioactivity = 0.3 & 3050 & 122550 & \bf 34520.0 \\
\hline
\end{tabular}
\end{center}
\caption{Results of GRADE algorithm for 5D task}
\label{rslt-B-grade}
\end{table}

\begin{figure*}[!ht]
\centering
\begin{tabular}{c@{\hspace{2mm}}c}
\fbox{\includegraphics*[width=6.5cm]{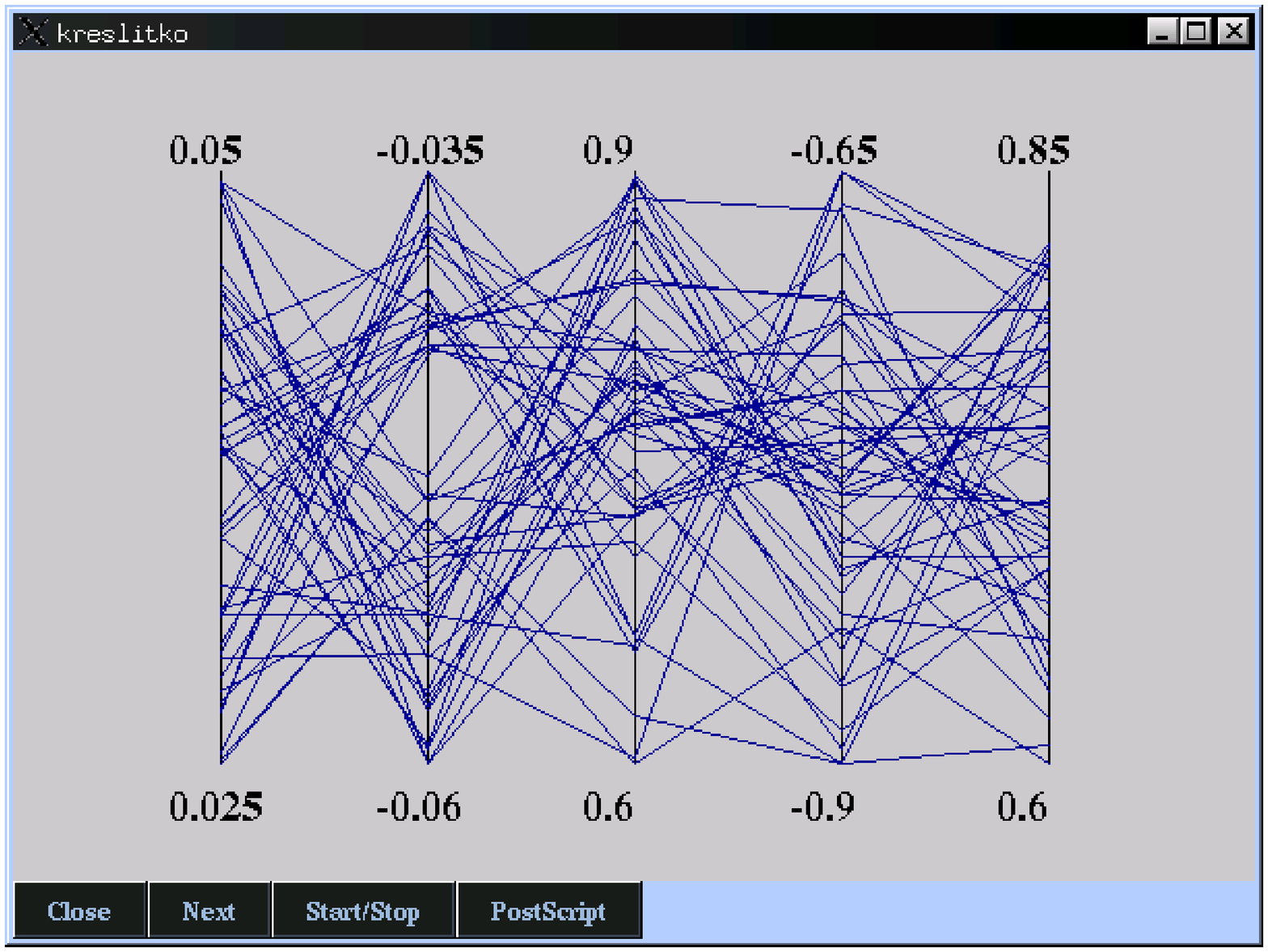}} &
\fbox{\includegraphics*[width=6.5cm]{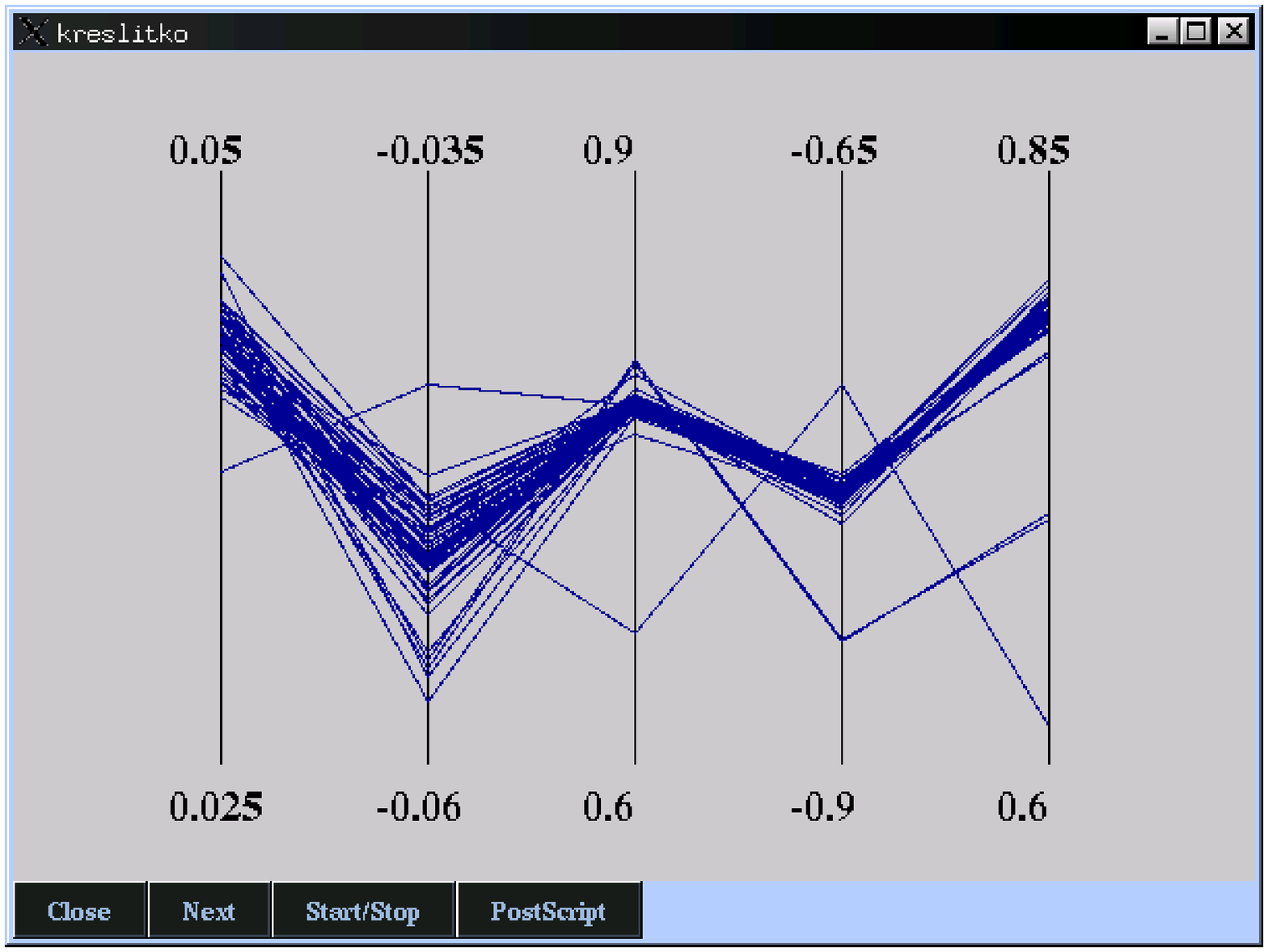}} \\
\\
G1:J=0.0091823 & G4:J=0.0071823 \\
\\
\fbox{\includegraphics*[width=6.5cm]{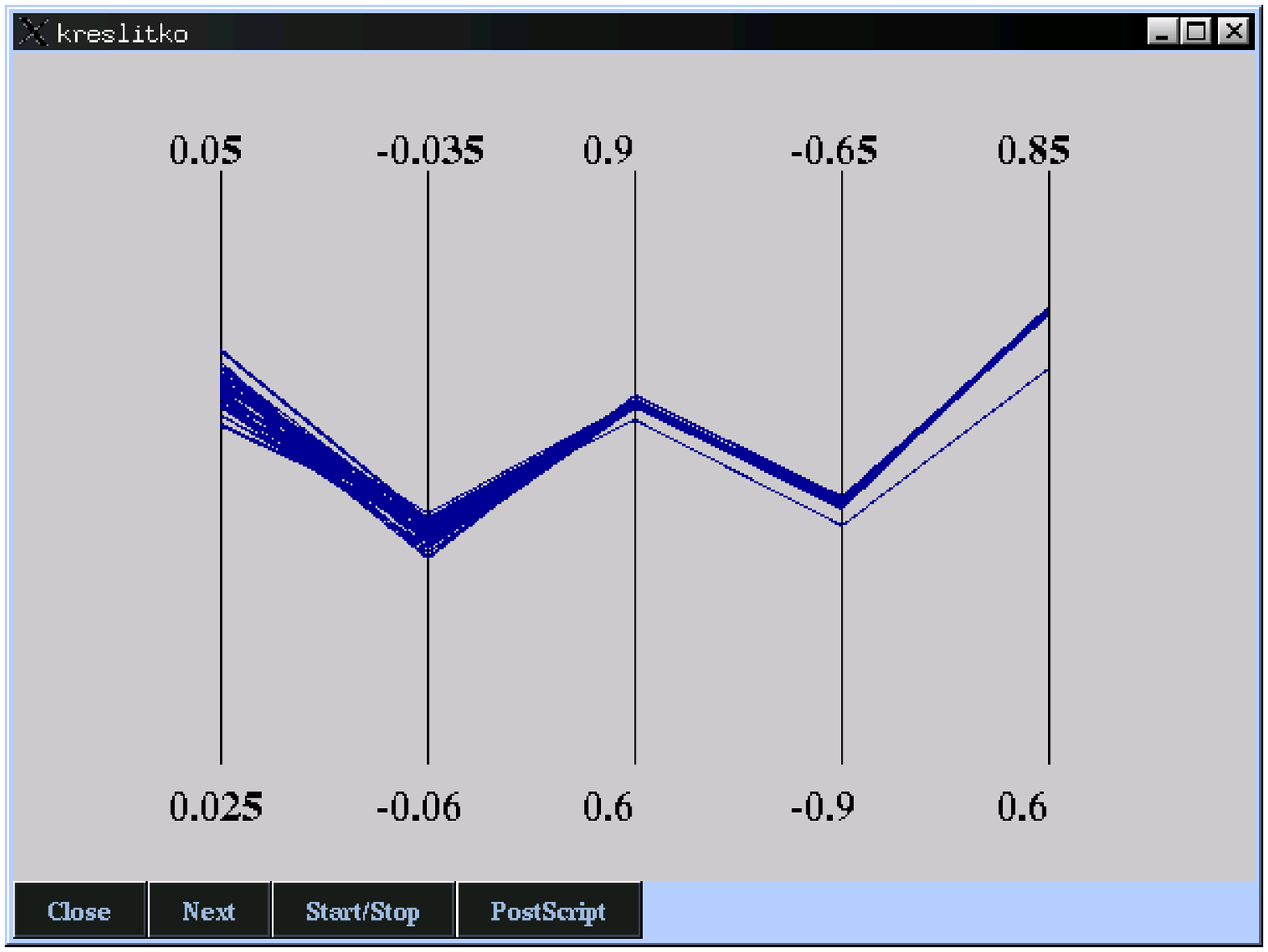}} &
\fbox{\includegraphics*[width=6.5cm]{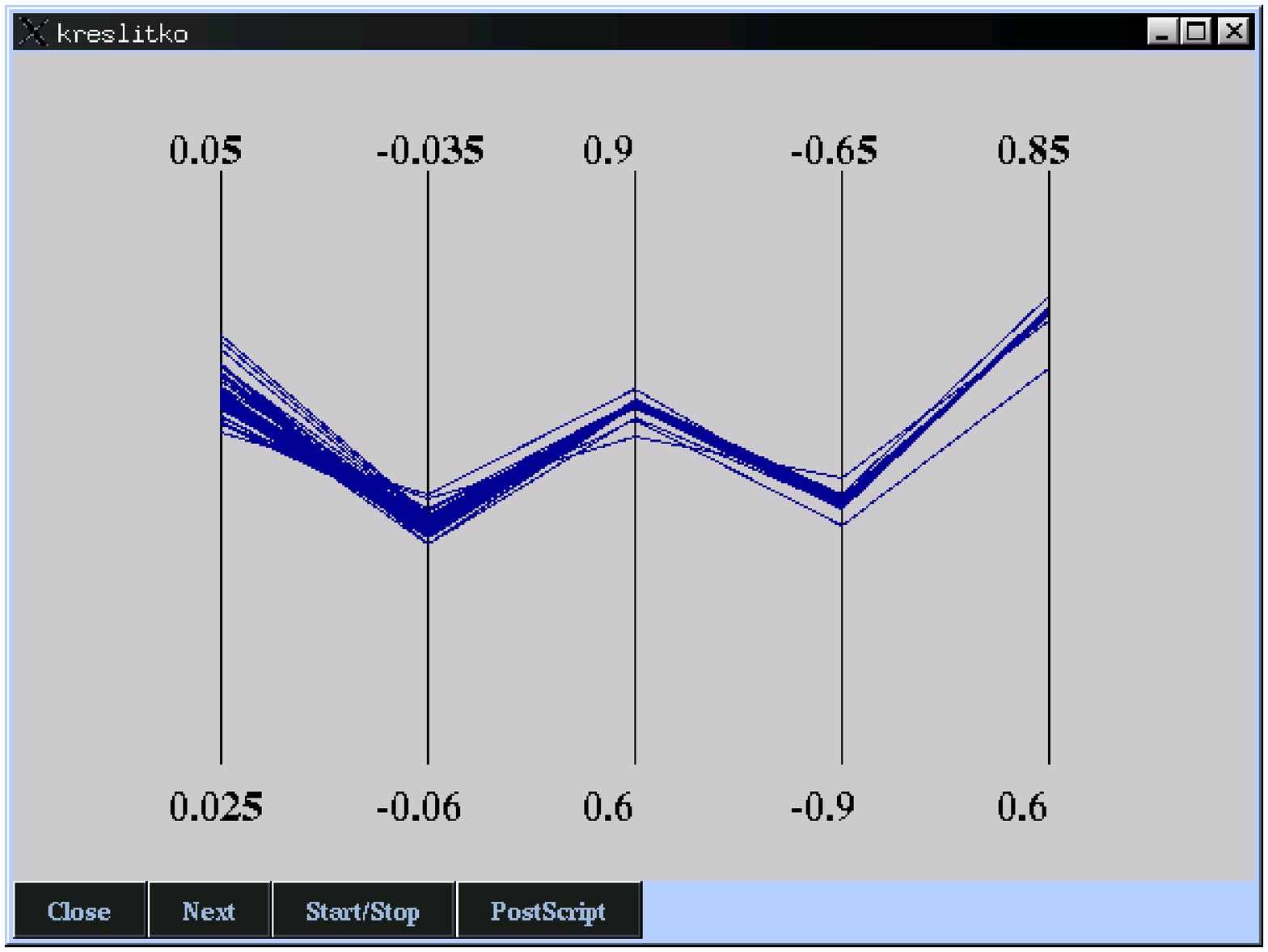}} \\
\\
G26:J=1.3302e-05 & G43:J=1.1839e-05 \\
\end{tabular}
\caption{Multibody system deployment: convergence of iterative
chromosome populations} \label{gartou}
\end{figure*}

One can notice the order of magnitude of increase in cost function
evaluation, which is brought about by a more elaborate form of the
cost function (see Figures \ref{becko2a} and \ref{becko2b}).
However, the latter is not the only reason. In this particular
problem the role of moments in the list of control variables is
much more important than the role of the horizontal and vertical
forces in bringing the system in the desired shape. This affects
the conditioning of the equations to be solved and the slow
convergence rate of the complete system is in reality only the
slow convergence of a single or a couple of control components.
The latter is illustrated in Figure \ref{gartou}, where we provide
the graphic representation of iterative values for computed
chromosomes, where every chromosome is represented by a continuous
line. We can note that the population of optimal values of moments
converges much more quickly than the force values which seek a
large number of iteration in order to stabilize. Another point
worthy of further exploration is the best way to accelerate the
convergence rate in the final computational phase.

\subsection{Optimal design of shear deformable cantilever}

\begin{figure}[htb]
\vspace{8mm}
\begin{center}
\includegraphics[width=100mm,keepaspectratio]{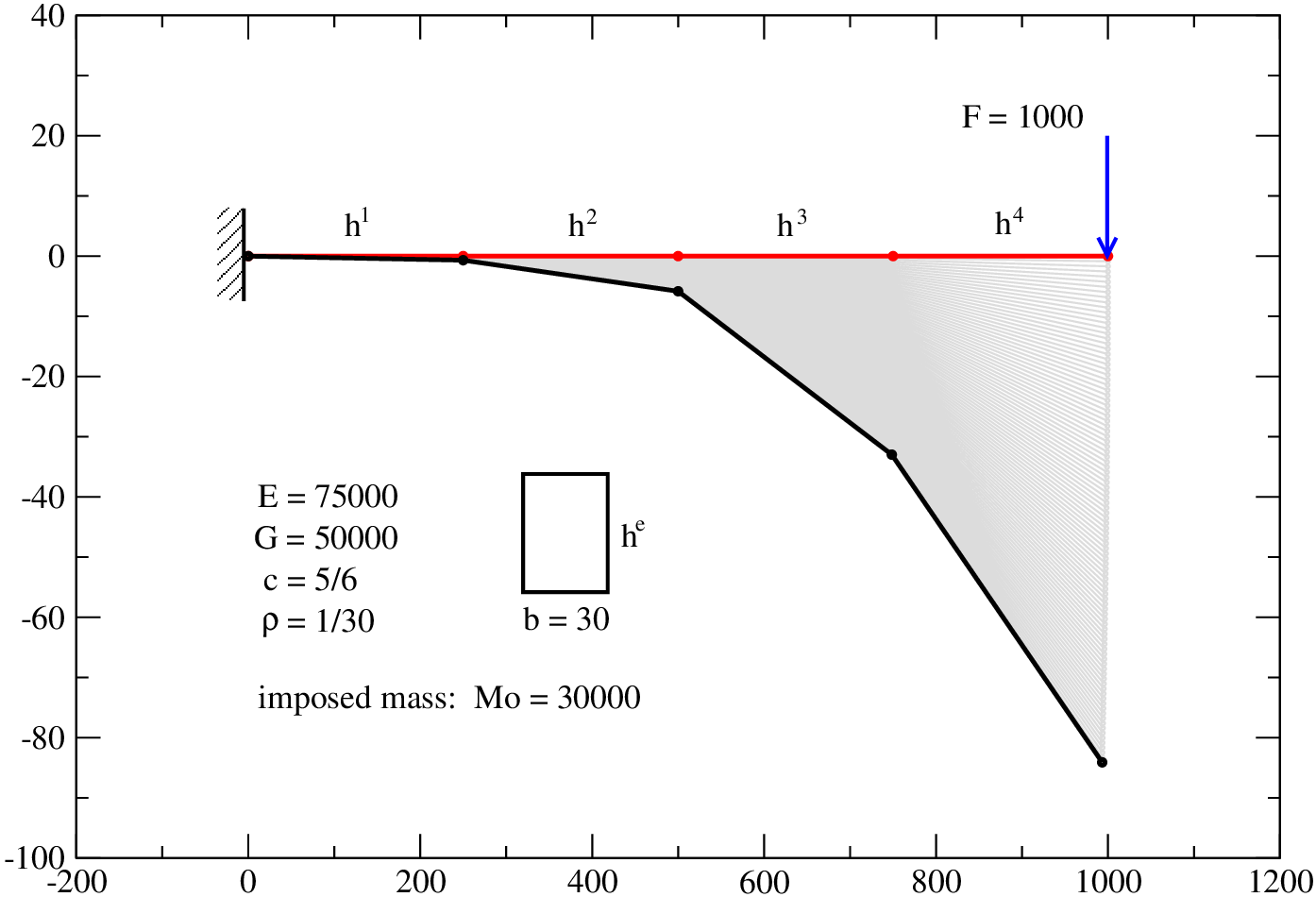}
\caption{Shear deformable cantilever beam optimal design : initial
and deformed shapes} \label{opt_fig}
\end{center}
\end{figure}

In this last example we study an optimal design problem which
considers the thickness optimization of a shear deformable
cantilever beam, shown in Figure \ref{opt_fig}.

The beam axis in the initial configuration of the cantilever and
the thickness is considered as the variable to be chosen in order
to assure the optimal design specified by a cost function. In the
setting of discrete approximation, we choose 4 beam elements each
with a constant thickness $h_i$, which results with 4 design
variables ${\bf d} \equiv {\bf h} = (h_1, h_2, h_3, h_4)$. The
beam mechanical and geometric properties are: Young's modulus $E =
75000$, shear modulus $G = 50000$, rectangular cross section $b
\times h_i$ with width $b = 30$ and mass density $\rho = 1/30$.
The latter is needed for computing the total mass of the beam $M =
\int_L \rho b h(s) \mbox{d}s$ to be used as the corresponding
limitation on the computed solution assuring reasonable values of
the optimal thickness under the free-end vertical force $F =
1000$. In order to assure a meaningful result the computations are
performed under chosen value of mass limitation is $M_0 = 30000$.
Other limitations are also placed on the admissible values of the
thickness for each element.

The first computation is performed by using the diffuse
approximation based response function and the sequential solution
procedure. The cost function is selected as the shear energy of
the beam and problem is cast as maximization of the shear energy,
with
\begin{equation}
J(\bmvarphi ({\bf d})) = \underset{G(\bmvarphi^{\ast}({\bf
d}^{\ast}), \cdot ) = 0}{\max} J(\hat{\bmvarphi}^{\ast}({\bf
d}^{\ast}))\ ; \quad J(\bmvarphi^{\ast}) = \int_L \frac{1}{2} GA
\gamma^2 \mbox{d}s
\end{equation}
where $\gamma$ is the shear strain component. The bounds on
thickness values are chosen as shown in Table \ref{tab_opt_hlim}.
\begin{table}[hbt]
\centering
\begin{tabular}{l|rrrr}
Thickness & $h_1$ & $h_2$ & $h_3$ & $h_4$\\
\hline
Min & 30 & 30 & 15 & 15 \\
Max & 60 & 60 & 35 & 35
\end{tabular}
\caption{Shear deformable cantilever optimal design : thickness
admissible values} \label{tab_opt_hlim}
\end{table}
The diffuse approximation computations on the grid are started
from an
  initial guesses for thickness ${\bf h}_0 = ( 55,
50, 30, 20 )$. It took 11 iterations to converge to solution given
as
$${\bf h} = ( 45.094, 40.074, 19.832, 15.000 )$$
The corresponding value of shear energy for this solution is
$J_{appr} = 16.3182$; we recall it is only an approximate
solution, since the computed value does not correspond to any of
the grid nodes.

The same solution is next sought by using GRADE algorithm; The
chosen values for GRADE parameters are: $PR = 10$, $CL = 1.0$ and
'radioactivity'$ = 0.2$. The genetic algorithm is executed 100
times, leading to the computational statistics reported in
Table~\ref{tab_opt_nsim0}.
\begin{table}[hbt]
\centering
\begin{tabular}{c|rrrc}
 & Minimum & Maximum & Mean Value  \\
\hline nb. of comput. $J(\cdot)$ & 120 & 3400 & 674.8
\end{tabular}
\caption{Shear deformable cantilever optimal design : computation
statistics} \label{tab_opt_nsim0}
\end{table}

The algorithm yielded two different solutions, both essentially
imposed by the chosen bounds; Namely, out of 100 runs,  57
converged to ${\bf h} = ( 60, 30, 15, 15 )$, with the
corresponding value of  $J = 17.974856$, whereas 43 settled for
${\bf h} = ( 30, 60, 15, 15 )$ with a very close value of $J =
17.926995$. Hence, each of two solutions leads to an improved
value of the cost function.

The second part of this example is a slightly modified version of
the first one, in the sense that the mechanics part of the problem
is kept the same and only a new cost function is defined seeking
to minimize the Euclidean norm of the computed displacement
vector, i.e.
\begin{equation}
J(\bmvarphi ({\bf d})) = \underset{G(\bmvarphi^{\ast}({\bf
d}^{\ast}), \cdot ) = 0}{\max} J(\hat{\bmvarphi}^{\ast}({\bf
d}^{\ast}))\ ; \quad J(\bmvarphi^{\ast}) = \frac{1}{2} \int_L \|
\bmvarphi - {\bf x} \|^2 \mbox{d}s
\end{equation}
Such a choice of the cost function is made for being well known to
result with a well-conditioned system expressing optimality
conditions. Indeed, the same type of sequential solution procedure
using diffuse approximation of cost function now needs only a few
iterations to find the converged solution, starting from a number
initial guesses. The final solution value is given as
$${\bf h} = (42.579, 35.480, 26.941, 15.000)$$.

In the final stage of this computation we recompute the solution
of this problem by using the genetic algorithm. The admissible
value of the last element thickness is also slightly modified by
reducing the lower bound to 5 (instead of 15) and higher bound to
25 (instead of 35) in order to avoid the optimal value which is
restricted by a bound. The first solution to this problem is
obtained by using again the sequential procedure, where the GRADE
genetic algorithm is employed at the last stage. The computed
value of the displacement vector norm for found solution is
$623808$ and mass $M$ is $30062$. The computations are carried out
a hundred times starting from random initial values. The
statistics of these computations are given in Table
\ref{tab_opt_nsim1}.

\begin{table}[hbt]
\centering
\begin{tabular}{c|rrrc}
 & Minimum & Maximum & Mean Value & Standard deviation\\
\hline
$h_1$ & 43.772 & 43.807 & 43.790 & 0.0094 \\
$h_2$ & 35.914 & 35.949 & 35.932 & 0.0088 \\
$h_3$ & 26.313 & 26.346 & 26.328 & 0.0082 \\
$h_4$ & 14.184 & 14.210 & 14.197 & 0.0064 \\
\hline nb. of comput. $J(\cdot)$ & 1440 & 9960 & 3497 &
--------
\end{tabular}
\caption{Shear deformable cantilever optimal design : computation
statistics} \label{tab_opt_nsim1}
\end{table}

The same kind of problem is now repeated by using the simultaneous
solution procedure, where all the optimality condition are treated
as equal and solved simultaneously resulting with 4 thickness
variables, 15 displacement and rotation components and as many
Lagrange multipliers as unknowns. The latter, in fact, is
eliminated prior to solution by making use of optimality condition
in (\ref{eq_4.13}$)_2$. The chosen upper and lower bounds of the
admissible interval are chosen as
\begin{equation}
\bmvarphi \in \left[ (1-EP)\bmvarphi^p, (1+EP)\bmvarphi^p \right]
\end{equation}
where the guess for the displacement $\bmvarphi^p$ is obtained by
solving mechanics problem with the values of thickness parameters
given in  Table \ref{tab_opt_nsim1}. The limitation on total mass
is added to the cost function. The choice of GRADE algorithm
parameters is given as $PR = 20, CL = 2$ and 'radioactivity' equal
to $0.1$. The computation is stopped with a fairly loose
tolerance, which allows to accelerate the algorithm convergence
but does not always lead a unique solution. Yet, the results in
Table \ref{tab_opt_sim1} show that standard deviation indeed
remains small, or that the solution is practically unique.

\begin{table}[hbt]
\centering
\begin{tabular}{c|rrrc}
 & Minimum & Maximum & Mean Value & Standard deviation\\
\hline
$h_1$ & 43.782 & 43.794 & 43.789 & 0.0026 \\
$h_2$ & 35.925 & 35.935 & 35.930 & 0.0021 \\
$h_3$ & 26.315 & 26.324 & 26.319 & 0.0019 \\
$h_4$ & 14.197 & 14.202 & 14.200 & 0.0010 \\
\hline nb. of comput. ${\bf r}^T{\bf r}$ & 111340 & 968240 &
313006 &
-------- \end{tabular}
\caption{Shear deformable cantilever optimal design : simultaneous
computation statistics} \label{tab_opt_sim1}
\end{table}


\section{Acknowledgements}
This work was supported by the French Ministry of Research and the European
Student Exchange Program ERASMUS. This support is gratefully acknowledged.

\section{Conclusions} \label{sec_conclusions}

The approach advocated herein for dealing with a coupled problem
of nonlinear structural mechanics and optimal design or optimal
control, which implies bringing all the optimality conditions at
the same level and treating all variables as independent rather
than considering equilibrium equations as a mere constraint and
state variables as dependent on design or control variables,  is
fairly unorthodox and rather unexplored. For a number of
applications the proposed approach can have a great potential. In
particular, the problems of interest to this work concern the
large displacements and rotations of a structural systems. The key
ingredient of such an approach pertains to geometrically exact
formulation of a nonlinear structural mechanics problem, which
makes dealing with nonlinearity description or devising the
solution schemes much easier then for any other model of this
kind. The model problem of the geometrically exact beam explored
in detail herein is not the only one available in this classl; We
refer to work in \cite{Ibrahimbegovic:1994} for shells or to
\cite{Ibrahimbegovic:1995} for 3D solids, with both models sharing
the same configuration space for mechanics variables as 3D beam.
The latter also allows to directly exploit the presented
formulation and the solution procedures of a coupled problem of
nonlinear mechanics, for either shells or 3D solids, and optimal
control or optimal design.

Two different solution procedures are presented herein; the first
one, which exploits the response surface representation of the
true cost function followed by a gradient type solution step,
leads to only an approximate solution. Although the quality of
such a solution can always be improved by further refining the
grid which serves to construct the response surface, the exact
solution is never computed unless the minimum corresponds to one
of the grid points. The second solution procedure, which solves
simultaneously the optimality conditions and nonlinear mechanics
equilibrium equations, does deliver the exact solution, although
often only after the appropriate care is taken to choose the
sufficiently close initial guess and to select the admissible
intervals of all variables accordingly. Probably the best method
in that sense is the combination of sequential and simultaneous
procedure, where the first serves to reduce as much as possible
the admissible interval and provide the best initial guess,
whereas the second furnishes the exact solution.

A number of further improvements can be made for the proposed
methods of this kind, to help increase both the convergence rates
and accuracy of the computed solution. One has to remember that
even only mechanics component equilibrium equations for a problem
of this kind can be very difficult to solve, since the large
difference in stiffness in different modes (e.g. bending versus
stretching) can result with a poorly conditioned set of equations.
Adding the optimality conditions on top only further increases the
difficulty. Finding the best possible way to deal with this
problem is certainly worthy of further explorations.

\bibliographystyle{plain}
\bibliography{list}
\end{document}